\pdfoutput=1

\documentclass[11pt]{article}

\usepackage[final]{acl}

\usepackage{times}
\usepackage{dsfont}
\usepackage{latexsym}
\usepackage{microtype}
\usepackage{amssymb}
\usepackage{tcolorbox}
\usepackage{amsmath}
\usepackage{fdsymbol}
\usepackage{bbm}
\usepackage{inconsolata}
\usepackage{float}
\usepackage{multirow}
\usepackage{graphicx}
\usepackage{booktabs}
\usepackage{caption}
\usepackage{subcaption}
\usepackage{bbold}
\usepackage{xspace}
\usepackage{makecell}
\usepackage{enumerate}

\usepackage[capitalize,noabbrev]{cleveref}
\usepackage{tcolorbox}
\usepackage[T1]{fontenc}

\usepackage{algorithm}
\usepackage{algpseudocode}
\usepackage{amsmath}
\usepackage{pifont}
\usepackage{longtable}

\usepackage[utf8]{inputenc}

\newcommand\eg{\emph{e.g.},\xspace}

\newcommand\ie{\emph{i.e.},\xspace}

\newcommand{\papertitle}{Legal Rule Induction\xspace}

\definecolor{stepcolor}{HTML}{d79b00}
\definecolor{contentcolor}{HTML}{6c8ebf}

\definecolor{cYellow}{RGB}{255,255,3}
\definecolor{cBlue}{RGB}{69,123,157}
\definecolor{cRed}{RGB}{231,56,71}
\definecolor{cRed_1}{RGB}{191,30,46}
\definecolor{cGray}{RGB}{168,218,219}
\definecolor{cBlue_2}{RGB}{5,48,97}
\definecolor{cBlue_1}{RGB}{115,186,214}
\definecolor{cBlue_3}{RGB}{13,76,109}
\definecolor{cBlue_4}{RGB}{64,121,160}
\definecolor{cOrange}{RGB}{250,134,0}
\definecolor{cBlue_6}{RGB}{13,76,109}
\definecolor{cBlue_7}{RGB}{16,106,130}
\definecolor{cBlue_8}{RGB}{19,136,160}
\definecolor{cBlue_9}{RGB}{115,184,214}
\title{\papertitle: Towards Generalizable Principle Discovery from Analogous Judicial Precedents}

\author{
Wei Fan\textsuperscript{1},
Tianshi Zheng\textsuperscript{1},
Yiran Hu\textsuperscript{2},
Zheye Deng\textsuperscript{1},
Weiqi Wang\textsuperscript{1},\\
\textbf{
Baixuan Xu\textsuperscript{1},
Chunyang Li\textsuperscript{1},
Haoran Li\textsuperscript{1}\thanks{Corresponding Authors},
Weixing Shen\textsuperscript{2}\footnotemark[1],
Yangqiu Song\textsuperscript{1}}\\
\textsuperscript{1}Department of Computer Science and Engineering, HKUST, Hong Kong SAR, China\\
\textsuperscript{2}School of Law, Tsinghua University, Beijing, China\\
\texttt{\{wfanag, hlibt\}@connect.ust.hk}, \texttt{wxshen@tsinghua.edu.cn}, \texttt{yqsong@cse.ust.hk}\\
}

\begin{document}
\maketitle

\begin{abstract}
Legal rules encompass not only codified statutes but also implicit adjudicatory principles derived from precedents that contain discretionary norms, social morality, and policy.
While computational legal research has advanced in applying established rules to cases, \emph{inducing} legal rules from judicial decisions remains understudied across jurisdictions.
The advent of Large Language Models~(LLMs) offers unprecedented potential to automate the extraction of such latent principles, yet progress is stymied by the absence of formal task definitions and benchmarks.
To address this gap, we formalize \textbf{Legal Rule Induction}~(LRI)\footnote{The LRI dataset is available at \url{https://github.com/HKUST-KnowComp/Legal-Rule-Induction}} as the task of deriving concise, generalizable doctrinal rules from analogous precedents, distilling their shared preconditions, normative behaviors, and legal consequences.
We further propose a reproducible pipeline for LRI dataset construction and, instantiating it on Chinese law as a representative jurisdiction, introduce the first LRI benchmark comprising 5,121 case sets (38,088 court cases) for model tuning and 216 expert-annotated gold test sets.
Experimental results reveal that:
1)~State-of-the-art LLMs struggle with over-generalization and hallucination;
2)~Training on our dataset markedly enhances LLMs’ capabilities in capturing nuanced rule patterns across similar cases.
\end{abstract}

\begin{quote}
\centering
\emph{"Common law courts have two functions: \\
resolving disputes according to\\
legal rules and making legal rules."} \\[1ex]
{\raggedleft\small\textemdash\ Melvin A. Eisenberg\par}

\end{quote}

\section{Introduction}
\begin{figure}[t!]
    \centering
    \includegraphics[width=0.40\textwidth]{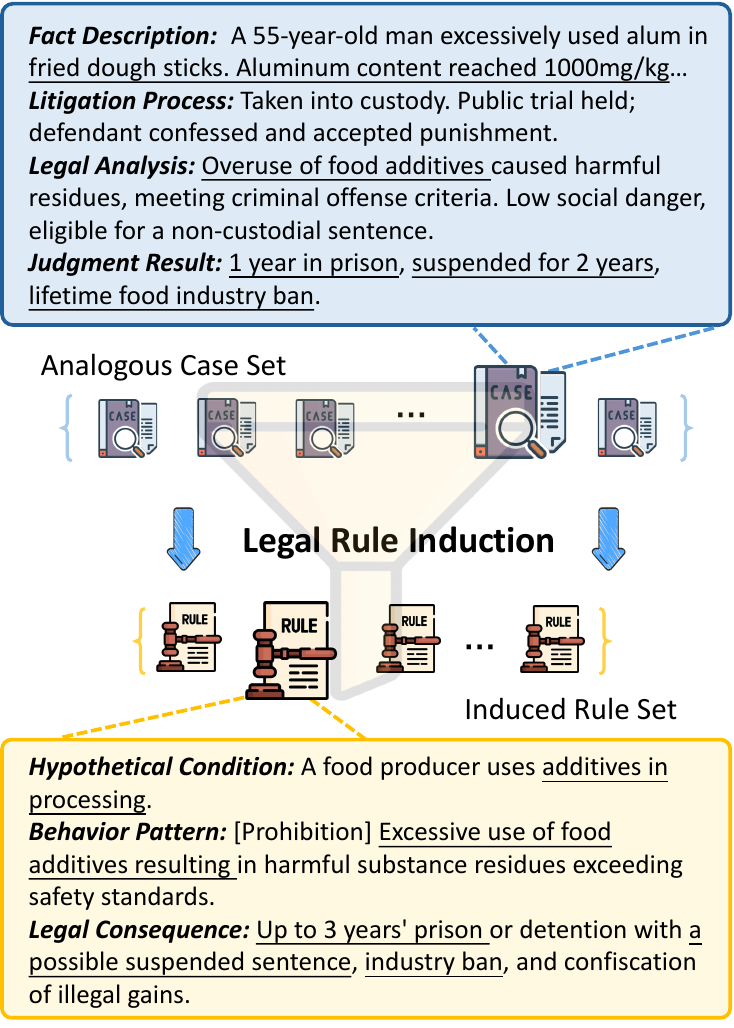}
    \caption{An illustration of legal rule induction from analogous judicial cases via the three-element logical structure of legal rules.}
    \label{figs:introduction}
\end{figure}

Modern legal systems, whether grounded in statutory codes or the case-law tradition, ultimately reason through legal rules~\cite{Eisenberg_2022}.
In civil law jurisdictions~(\eg China and France)~\cite{merryman2018civil, watkin2017historical}, rules are codified in statutory provisions characterized by explicit logical structures.
Common law systems, while also referencing statutes, primarily operationalize rules through precedent~\cite{holmes2020common}.
Under stare decisis~\cite{douglas1949stare}, a court is obliged to apply the rule articulated in any binding precedent, whether issued by a higher court or by itself, whenever the present case is materially indistinguishable~\cite{Eisenberg_2022}.
Although these systems differ superficially, explicit code articles versus implicit precedent rule~\cite{brewer2013precedents,Lamond_2005}, civil and common law rely on the same normative atom: the legal rule~\cite{dickinson1931legal}. Hence, the capacity to extract, articulate, and employ that atom is indispensable to any form of legal reasoning~\cite{levi2013introduction, guha2023legalbenchcollaborativelybuiltbenchmark}.

Current computational legal research tends to bifurcate statutory and precedent-based reasoning, framing the former as \textbf{deductive reasoning}~\cite{blairstanek2023gpt3performstatutoryreasoning} and the latter as similarity matching~\cite{liu2025improvingsimilarcaseretrieval}, neglecting their common grounding in rules.
This leaves legal \textbf{inductive reasoning}~(\ie rule induction), a cornerstone of everyday legal work, critically underexplored.
As Melvin Eisenberg highlights, common law courts not only resolve disputes by applying established rules but also formulate new rules from clusters of earlier decisions~\cite{Eisenberg_2022}.
Inductive reasoning is equally vital in civil law systems: China's Supreme People's Court mandated similar-case retrieval during adjudication\footnote{\url{https://www.court.gov.cn/zixun/xiangqing/243991.html}}, requiring judges to identify shared patterns across analogous cases, which inherently relies on inductive generalization.
In practice, lawyers, litigants, and judges across jurisdictions spend considerable effort extracting abstract propositions from massive corpora of judgments.
The advent of LLMs~\cite{deepseekai2025deepseekv3technicalreport,openai2024gpt4technicalreport,qwen2025qwen25technicalreport}, with extensive context windows and strong reasoning capabilities, raises the possibility of automating this process. Yet across jurisdictions, the task remains under-defined: there is no precise task definition, no public benchmark, and no standard methodology.

To bridge this gap, we formally propose the \textbf{L}egal \textbf{R}ule \textbf{I}nduction~\textbf{(LRI)} task, a jurisdiction-agnostic formulation for synthesizing abstract legal rules from analogous judicial precedents, as illustrated in~\cref{figs:introduction}. Informed by jurisprudence in China, we define a legal rule by three core elements: hypothetical applicability conditions triggering the rule, behavioral prescriptions that govern conduct (permitting, prohibiting, or obligating actions), and legal consequences specifying outcomes, whether positive (\eg rights conferred) or negative (\eg punishments imposed). Input precedents for the LRI task consist of facts, procedural history, legal analysis, and judgment, excluding statutory citations to compel models towards genuine rule induction rather than mere recall of codified law~\cite{louis2023findinglawenhancingstatutory}.

To facilitate LRI research and benchmark LLM performance, we introduce the \textbf{LRI Dataset}, constructed through a reproducible pipeline applicable across jurisdictions.
The core pipeline~(collecting judgments, clustering by shared legal grounds, and extracting rules) is jurisdiction-agnostic and only the data source varies by legal system.
In common law systems, legal rules are derived from both statutory citations and reasoning embedded in precedents, making implicit rule extraction labor-intensive.
Civil law judgments, by contrast, cite and apply codified statutes directly, enabling more straightforward case-to-rule alignment.
Exploiting this feature, we focus on \textit{China Judgments Online}\footnote{\url{https://wenshu.court.gov.cn/}}, one of the largest and most authoritative data sources in civil law jurisdictions, from which we collect over 9 million civil and criminal cases and cluster them into case sets referencing the same statutory articles.
Each set thereby shares explicit statutory grounding while also revealing implicit discretionary principles through the courts’ analyses. 
Following an automated processing pipeline via DeepSeek-R1~\cite{deepseekai2025deepseekr1incentivizingreasoningcapability} and applying filters based on set size and rule applicability, we curate the \textbf{LRI-AUTO} dataset of 5,121 case sets (38,088 judgments) for model tuning.

For evaluation, we further develop \textbf{LRI-GOLD}, a curated test set of 216 case sets (1,620 cases) annotated by legal researchers.
We evaluate a range of leading LLMs, including foundational and reasoning models~\cite{xu2025largereasoningmodelssurvey}, as well as an iterative induction-verification pipeline. Results reveal persistent challenges such as hallucination and overgeneralization, yet confirm measurable progress.
Notably, smaller-scale LLMs~(3B-8B parameters) tuned on our dataset achieve over 76\% gains in F1-scores, outperforming larger closed-source models.
While our benchmark centers on Chinese law, the LRI task construction pipeline is jurisdiction-agnostic, offering a replicable foundation for future research across diverse legal systems.








\section{Related Work}
\subsection{Legal Reasoning in Computational Law}
In computational law, research on legal reasoning spans several paradigms: Legal Document Summarization~\cite{zhong2022computing, shen2022multi, polsley2016casesummarizer}, Legal Argument Mining~\cite{santin2023argumentation, poudyal2020echr, palau2009argumentation}, Legal Question Answering~\cite{zhang2023glqa, sovrano2020legal, louis2024interpretable}, and Legal Judgment Prediction~\cite{zhong2020iteratively, zhang2023contrastive, chalkidis2019neural}.
Advances in LLMs~\cite{minaee2025largelanguagemodelssurvey} further extend these capabilities to applications such as legal consultation~\cite{cui2023chatlaw}, contract review~\cite{graham2023natural}, and drafting~\cite{wang2025acord}.
Earlier work in legal technology has also investigated rule extraction from regulatory texts~\cite{dragoni2017combining, Wyner2011OnRE, francesconi2009approach}, relying on symbolic or semantic pattern matching rather than inductive generalization. While such methods contributed annotated datasets on Italian, Australian, and other national regulations, they operated at the level of logical representations rather than natural language.
In practice, due to the complexity and the incompatibility of legal knowledge across systems, legal reasoning tasks are conventionally confined to a single jurisdiction.
The advent of LLMs enables natural language rule formation, allowing models to infer and generalize rules from precedents.

\subsection{Inductive Reasoning}
Inductive reasoning~\cite{heit2000properties} is a fundamental cognitive process that involves drawing general conclusions from specific observations.
Cognitive science frames induction as probabilistic belief revision under the Bayesian framework~\cite{tenenbaum2011grow}, where learning arises from combining prior knowledge with observed data to derive posterior probabilities~\cite{lake2015human}.
NLP research in inductive reasoning recently shifts from task-specific architectures~\cite{odena2020bustle,tian2020learning,sable2022language} to large pre-trained models capable of broad inductive inference in natural language~\cite{yang2022language,mirchandani2023large,gendron2023large}.
LLMs equipped with extremely long context windows \emph{(> 100k tokens)} and \emph{thinking} ability~\cite{wei2022chain, deepseekai2025deepseekr1incentivizingreasoningcapability} can ingest multiple full-length cases and surface latent regularities without manual feature engineering.  Consequently, legal rule discovery is evolving from static symbol manipulation to dynamic pattern extraction in free text. Crucially, this evolution provides systematic evidence against critiques positing legal reasoning as fundamentally analogy-based~\cite{sherwin1999defense} or similarity-based~\cite{schauer1987precedent}.

\section{Preliminaries}
\subsection{Task Definition}
We define Legal Rule Induction~(LRI) as the task of algorithmically deriving a concise set of normative rules from a given collection of precedent cases.
Formally, given a \textbf{precedent case set} $\mathcal{P} = \{p_i\}_{i=1}^M$ where $M \in \mathbb{N}^+$ ranges between 5 and 10 inclusive. Each case $p_i$ within this set is a structured entity comprising four key components: a \textit{fact description}, the \textit{litigation process}, a \textit{legal analysis}, and the final \textit{judgment result}.
The objective is to induce a rule set $\mathcal{R} = \{r_j\}_{j=1}^N$. Each rule $r_j$ in this set comprises three elements: a \textit{hypothetical condition}, a \textit{behavior pattern}, and a \textit{legal consequence}.
To ensure that each induced rule reflects a generalizable normative pattern, rather than an isolated exception, we adopt a principle from frequent pattern mining~\cite{10.5555/645920.672836}. Specifically, each rule $r_j$ must apply to a strict majority of the precedent cases in $\mathcal{P}$.
We define the support set of a rule $r_j$, denoted $\text{Supp}(r_j, \mathcal{P})$, as:
\begin{equation}
    \text{Supp}(r_j, \mathcal{P}) = \{ p_i \in \mathcal{P} \mid r_j \text{ applies to } p_i \}.
\end{equation}
Accordingly, the coverage condition for each rule is:
\begin{equation}
    |\text{Supp}(r, \mathcal{P})| > \frac{M}{2}.
\end{equation}

\subsection{Legal Rule}
To ensure that the LRI task aligns with well-established legal reasoning and facilitates cross-cultural generalizability, we consider legal rules along three conceptual dimensions.
From the perspective of doctrinal scope, this study focuses on \textbf{criminal law}, \textbf{civil law}, and their associated \textbf{procedural laws}~\cite{dong2023procedural}. Legal domains that are highly specialized or jurisdictionally narrow, such as administrative regulations or municipal by-laws, are excluded from consideration.
From the perspective of normative function, each induced rule is required to instantiate one of the foundational deontic modalities recognized in legal theory: \textbf{permission} (an action is allowed), \textbf{prohibition} (an action is forbidden), or \textbf{obligation} (an action is required). More complex or context-dependent normative forms, such as conditional obligations or defeasible rules, fall outside the scope of the present formulation and are left for future work.
From the perspective of normative source, legal rules are divided into two broad categories. (1) \textbf{Explicit rules} are those that are codified in statutes, regulations, or other formal legal texts. (2) \textbf{Implicit rules}, by contrast, are not directly articulated in legal provisions but are inferred through analysis of legal reasoning across precedent cases. For example, in Chinese criminal adjudication, male defendants often receive harsher sentences than female defendants for comparable offenses, even though no statute explicitly mandates such a distinction. Research in criminology has identified a range of such extra-legal factors, including defendant demographics, case context, and judicial attitudes, that systematically influence sentencing outcomes~\cite{Ulmer19092023, https://doi.org/10.1111/j.1745-9125.2004.tb00516.x, ijcai2024p833}. We define the normative patterns by which these extra-legal factors shape legal consequences as implicit rules.

\begin{figure*}[t]
    \centering
    \includegraphics[width=2\columnwidth]{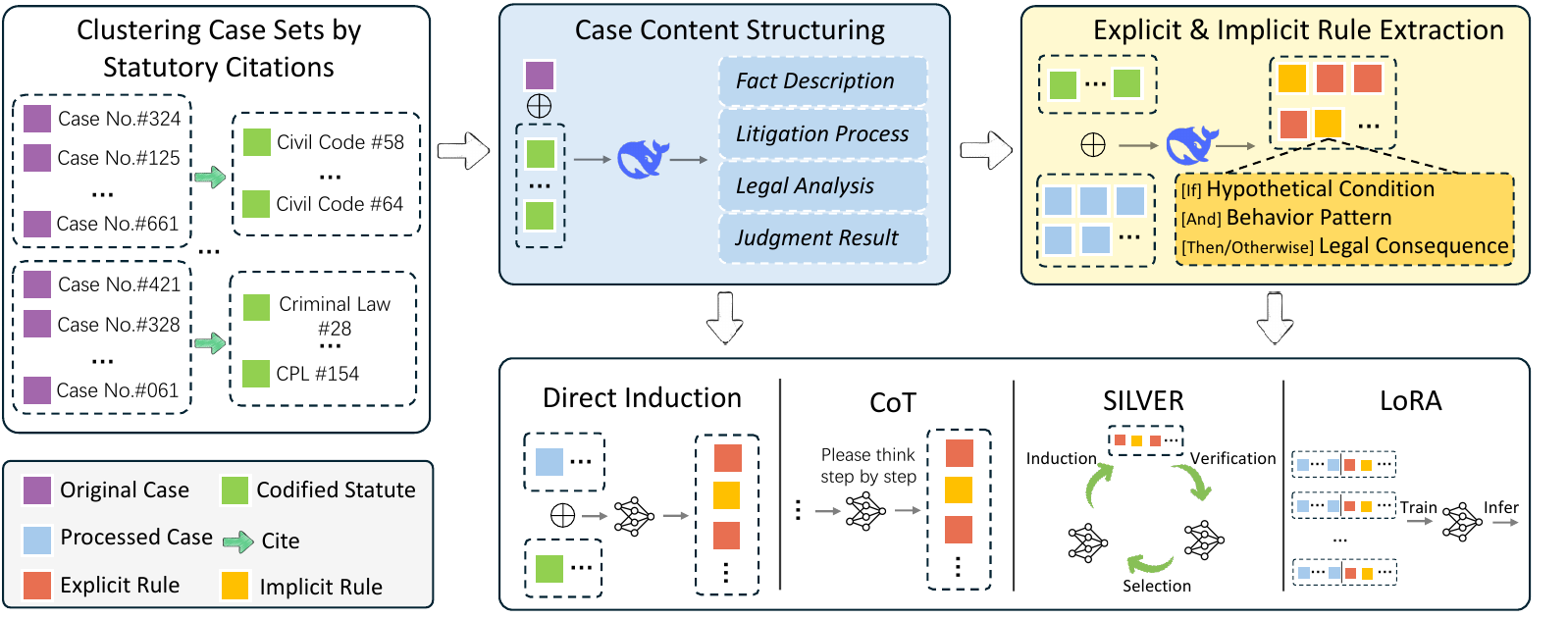}
    \caption{The overview of the \textbf{LRI-AUTO} dataset curation pipeline (for civil and criminal cases) and main methods for rule induction, including LoRA, which utilizes LRI-AUTO for tuning and the \textbf{LRI-GOLD} dataset for testing.}
    \label{figs:method}
\end{figure*}

\subsection{Inductive Reasoning Pipeline}\label{pre:pipeline}
In the main experiments, we consider four training-free pipelines in inductive reasoning:
\paragraph{Direct Induction}
This pipeline employs LLMs to generate normative rules directly from the provided case texts using a single-step prompting strategy.
Following~\cite{zheng2025logidynamicsunravelingdynamicslogical}, we consider the direct output of LLMs in this manner as a form of baseline inductive inference.

\paragraph{Chain-of-Thought~(CoT)}
CoT prompting~\cite{wei2022chain} operationalizes a more deliberative, multi-step reasoning process. It guides the LLM to decompose the rule induction task into intermediate analytical stages (\eg identifying common factual patterns, discerning judicial reasoning, and then formulating a rule).

\paragraph{Long Chain-of-Thought}\label{pre:LRM}
Long-CoT refers to the phenomenon where Large Reasoning Models (LRMs), such as o1~\cite{jaech2024openai} and DeepSeek-R1~\cite{deepseekai2025deepseekr1incentivizingreasoningcapability}, spontaneously generate extended chains of reasoning before answering complex questions.

\paragraph{SILVER}
To further advance rule induction for LRI, we propose \textbf{SI}mp\textbf{L}y Iterative Induction and \textbf{VER}ification~\textbf{(SILVER)}, which implements an induction–verify–update loop~\cite{qiu2024phenomenalpuzzlingtestinginductive}. 
As detailed in~\cref{alg:silver}, the process commences with an initial set of rules induced from the case sets.
Subsequently, SILVER alternates between another two core stages as detailed in~\cref{app:prompts_silver}: (i) verifying each candidate rule against the case set to determine if it surpasses the predefined majority-support threshold, and (ii) re-inducing fresh candidate rules to address aspects of the cases not adequately covered by the already verified ones.
This cycle repeats until no new high-support rules are found or a maximum iteration count is reached.





\section{Legal Rule Induction Dataset}
In this section, we present the Legal Rule Induction dataset curation pipeline, as detailed in~\cref{figs:method}, and provide dataset statistics.

\subsection{Corpus and Clustering}\label{sec:corpus_clustering}
Chinese legal cases typically specify cited legal article numbers, enabling large-scale automated clustering of case sets sharing common legal bases.
We collect over 9 million criminal/civil cases from China Judgments Online (CJO) and their contemporaneous legal provisions to ensure citation consistency (see~\cref{app:cjo}).
Using regex, we extract all legally cited provisions of these cases from four core Chinese legal codes: the \textbf{\textit{Criminal Law}}, the \textbf{\textit{Civil Code}}, the \textbf{\textit{Criminal Procedure Law}}, and the \textbf{\textit{Civil Procedure Law}}. Then, cases citing identical legal provisions are automatically clustered into the same case sets, and the set size distribution is depicted in~\cref{figs:set_size_distribution}.

\subsection{Case Content Structuring}
Original documents contain regional formatting inconsistencies and sensitive information such as court names, personal identifiable information (PII), and legal article texts.
To isolate the core case content and legal citation for each case $p \in \mathcal{P}$, we employ the DeepSeek-R1 model~\cite{deepseekai2025deepseekr1incentivizingreasoningcapability} for content structuring with anonymisation.
Building on~\cite{huang-etal-2024-cmdl}, we identify and extract four key components from the court documents for each case.
We replace their \textit{relevant law} section from the structured case content with the litigation process (also known as procedural history) to avoid exposing legal articles/charges directly while ensuring LLMs access complete procedural context during rule induction.
Sensitive data (\eg names → ``Defendant A'', locations → ``City C'') is anonymised with generic substitutes, preserving demographic details (age/gender/occupation) where pertinent. Full implementation protocols are in~\cref{app:prompts_ccs}.

\begin{table}[t]
\centering
\small
\resizebox{0.95\linewidth}{!}{%
\begin{tabular}{lccc}
\toprule
& \textbf{\# Train} & \textbf{\# Test} & \textbf{\# Gold} \\

\midrule
\textit{Case Sets} & 4,552 & 569 & 216 \\
\textit{Civil Case Sets} & 2,847 & 347 & 108 \\
\textit{Criminal Case Sets} & 1,705 & 222 & 108 \\
\midrule
\textit{Cases} & 33,797 & 4,291 & 1,620 \\
\textit{Civil Cases} & 21,068 & 2,601 & 810 \\
\textit{Criminal Cases} & 12,729 & 1,690 & 810 \\
\midrule
\textit{Rules} & 26,372 & 3,278 & 1,132 \\
\textit{Explicit Rules} & 15,608 & 1,933 & 711 \\
\textit{Implicit Rules} & 10,764 & 1,345 & 421 \\
\midrule
\textit{Avg Case Length} & 569.5 & 567.1 & 569.0 \\
\textit{Avg Rule Per Case Set} & 5.79 & 5.76 & 5.24 \\
\midrule
\textit{Annotation} & \textbf{-} & \textbf{-} & \ding{52}\\
\bottomrule
\end{tabular}
} 
\caption{Statistics for automatically constructed LRI-AUTO~(Train / Test) and expert-annotated~LRI-GOLD.}
\label{tabs:dataset_stats}
\end{table}
\begin{figure}[t]
    \centering
    \includegraphics[width=0.47\textwidth]{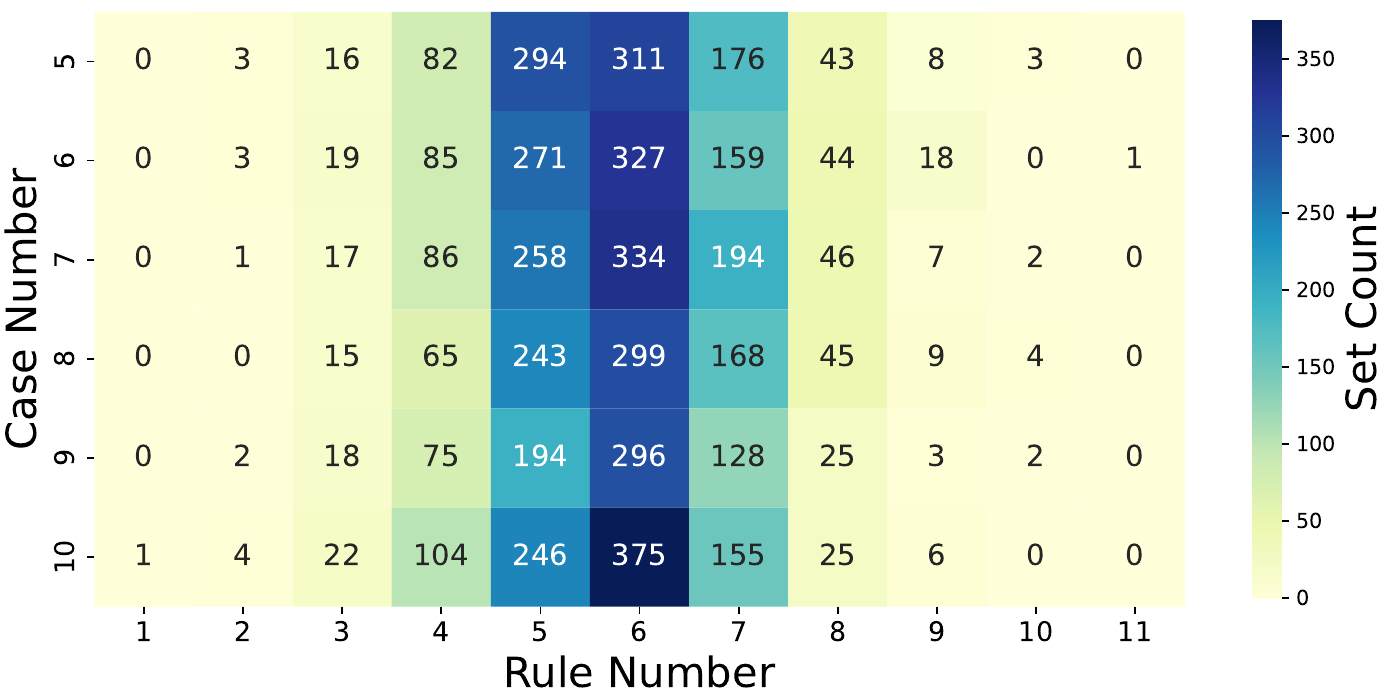}
    \caption{Distribution of rule set sizes across case numbers in the LRI Dataset.}
    \label{figs:heatmap}
\end{figure}

\subsection{Explicit and Implicit Rule Extraction}\label{sec:rule_extraction}
Legal provisions $\mathcal{S}_p$, defined as the set of statutory articles cited by the case set $\mathcal{P}$, associated with each case set $\mathcal{P}$ (~\cref{sec:corpus_clustering}), are unsuitable as direct ground truth rules for LRI.
Firstly, $\mathcal{S}_p$ often contains specific charges or offence names, skewing LRI towards statutory retrieval instead of rule induction. Secondly, cases may not use all parts of cited provisions, as articles often have multiple sub-clauses (\eg a case set might only pertain to one paragraph of a multi-paragraph article like Article 1079, PRC Civil Code, despite the entire article being cited).
Therefore, for each case set $\mathcal{P}$, DeepSeek-R1 is used to derive the two rule categories:
(1) \textbf{Explicit rules} $r_{\text{exp}}$: Rules directly from $\mathcal{S}_p$ applicable to all cases in $\mathcal{P}$, excluding specific charges/offense names.
(2) \textbf{Implicit rules} $r_{\text{imp}}$: Rules reflecting judicial practices or societal norms, not explicit in $\mathcal{S}_p$.
Rule extraction prompts and methodologies are detailed in~\cref{app:prompts_re}.

\begin{table*}[!t]
\renewcommand\arraystretch{1.00}
\small
\begin{center}
\begin{tabular}{
    m{2.5cm}|m{2.7cm}|
    m{1.2cm}<{\centering}m{1.2cm}<{\centering}|
    m{1.2cm}<{\centering}m{1.2cm}<{\centering}|
    m{1.2cm}<{\centering}m{1.2cm}<{\centering}
}
\toprule
\multirow{2}{*}{\textbf{Method}} & \multirow{2}{*}{\textbf{Model}} & 
\multicolumn{2}{c|}{\textbf{Rule Type}} & 
\multicolumn{2}{c|}{\textbf{Rule Level}} & 
\multicolumn{2}{c}{\textbf{Set Level}} \\
& & \textbf{Exp-Rec} & \textbf{Imp-Rec} & 
\textbf{Mic-Pre} & \textbf{Mic-F1} & 
\textbf{Mac-Pre} & \textbf{Mac-F1} \\
\midrule

\multirow{10}{*}{\makecell[l]{\textbf{LLMs} \textbf{(Direct)}}}
& GPT-4o-mini   & 45.99 & 29.22 & 57.25 & 46.92 & 58.41 & 46.86 \\
& GPT-4o    & 55.56 & 27.79 & \textbf{71.81} & 55.50 & \textbf{72.65} & 54.53 \\
& Gemini-2.5-Flash     & \underline{73.00} & 37.77 & 61.14 & 60.51 & 60.74 & 58.96 \\
& Llama-4-Scout & 47.26 & 25.18 & 58.47 & 46.82 & 60.87 & 45.85 \\
& Llama-4-Maverick   & 48.10 & 23.04 & 60.39 & 47.23 & 59.68 & 45.59 \\
& Qwen-2.5-72b   & 62.17 & 42.76 & 58.24 & 56.55 & 60.10 & 55.50 \\
& Qwen-Max   & 60.76 & \underline{43.47} & 61.32 & 57.61 & 60.05 & 56.14 \\
& DeepSeek-V3-0324   & 66.10 & \textbf{47.74} & 62.83 & \underline{61.00} & 61.66 & \underline{59.27} \\
& Claude-3.5-Sonnet   & 59.63 & 35.39 & 70.74 & 59.01 & \underline{70.73} & 58.40 \\
& Claude-3.7-Sonnet   & \textbf{74.68} & 42.99 & \underline{70.92} & \textbf{66.67} & 70.23 & \textbf{65.22} \\
\midrule

\multirow{10}{*}{\makecell[l]{\textbf{LLMs} \textbf{(CoT)}}}
& GPT-4o-mini & 41.49 & 15.68 & 67.98 & 43.42 & 67.69 & 42.53 \\
& GPT-4o & 41.63 & 14.49 & \textbf{80.95} & 45.39 & \textbf{78.36} & 43.72 \\
& Gemini-2.5-Flash & \underline{68.21} & 28.50 & 73.78 & \underline{61.99} & 74.14 & \underline{60.51} \\
& Llama-4-Scout & 45.85 & 17.10 & 71.97 & 47.24 & 75.25 & 46.41 \\
& Llama-4-Maverick & 41.49 & 14.73 & 72.71 & 43.99 & 72.03 & 41.90 \\
& Qwen-2.5-72b & 44.02 & 21.14 & 68.37 & 46.74 & 70.26 & 44.32 \\
& Qwen-Max & 54.29 & 24.47 & 72.44 & 54.12 & 72.76 & 52.95 \\
& DeepSeek-V3-0324 & 61.88 & \underline{32.07} & 69.70 & 58.76 & 71.87 & 56.65 \\
& Claude-3.5-Sonnet & 54.15 & 26.60 & 74.18 & 55.16 & 73.80 & 54.69 \\
& Claude-3.7-Sonnet & \textbf{70.89} & \textbf{37.77} & \underline{75.77} & \textbf{66.07} & \underline{76.66} & \textbf{65.17} \\
\midrule

\multirow{5}{*}{\makecell[l]{\textbf{LRMs} \\\textbf{(Long-CoT)}}}
& o3-mini & 46.41 & 13.78 & \textbf{83.08} & 48.53 & \textbf{84.04} & 48.76 \\
& Gemini-2.5-Flash & \underline{72.01} & 31.83 & 70.22 & 62.96 & 70.58 & 61.37 \\
& Deepseek-R1 & 62.87 & \underline{41.57} & \underline{74.31} & \underline{63.18} & \underline{74.45} & \underline{61.38} \\
& Claude-3.7-Sonnet & \textbf{75.39} & \textbf{43.94} & 68.02 & \textbf{65.78} & 67.94 & \textbf{64.33} \\
& Grok-3-mini & 42.76 & 13.30 & 70.73 & 43.88 & 71.18 & 43.94 \\
\midrule

\multirow{5}{*}{\makecell[l]{\textbf{LLMs} \\\textbf{(SILVER)}}}
& GPT-4o-mini & 68.92 & 38.48 & 58.01 & 57.80 & 55.13 & 54.95 \\
& Gemini-2.5-Flash & \textbf{88.19} & 43.71 & \underline{62.15} & \textbf{66.56} & \underline{60.42} & \textbf{63.82} \\
& Llama-4-Scout & 64.14 & 29.93 & \textbf{63.89} & 55.70 & \textbf{63.19} & 54.87\\
& Qwen-2.5-72b & 81.01 & \underline{52.02} & 57.11 & 63.00 & 52.68 & 57.82 \\
& DeepSeek-V3-0324 & \underline{84.81} & \textbf{64.77} & 56.99 & \underline{64.73} & 51.88 & \underline{59.90} \\

\bottomrule
\end{tabular}
\end{center}
\caption{Performance~(\%) on the LRI-GOLD benchmark across four baselines. \textbf{Exp-Rec} and \textbf{Imp-Rec} denote Micro Recall on explicit and implicit rules. We \textbf{bold} the best and \underline{underline} the second-best results in each baseline.}
\label{tabs:evaluation_overall_performance}
\end{table*}

\subsection{Case Set Postprocessing}
\paragraph{Rule Element Integrity Filter}
To ensure rule completeness, case sets are filtered if their corresponding rules, as extracted by DeepSeek-R1, lack essential elements in the \textbf{\textit{hypothetical condition}}, \textbf{\textit{behavior pattern}} (including action type), or \textbf{\textit{legal consequence}}.
This addresses potential omissions due to DeepSeek-R1 limitations, like hallucination or inconsistent instruction following.

\paragraph{Rule Applicability Filter}
A filtering step is applied to refine the rule sets: explicit rules $r_{\text{exp}}$ are retained only if they demonstrate 100\% applicability across all cases within their respective set $\mathcal{P}$. Implicit rules $r_{\text{imp}}$ are retained only if their applicability, as initially assessed, exceeds the 50\% threshold within their set.

\paragraph{Set Size Filter}
To manage the solution space for rule induction and constrain model input context, sets are filtered to retain those with over 5 cases. Sets exceeding 10 cases are randomly sampled down to 10. This results in final case sets $\mathcal{P}=\{p_1,p_2,\ldots,p_M\}$ containing 5 to 10 cases.

\subsection{LRI Dataset Collection and Annotation}
Following DeepSeek-R1 response collection and several filters, the LRI-AUTO dataset is constructed for model training.
This involves uniformly sampling approximately 1,000 instances from case set collections, categorized by the number of cases per set (ranging from 5 to 10). Each sampled instance comprises a case set $\mathcal{P}$ and its corresponding rule set $\mathcal{R}$.
For robust evaluation, we further construct the LRI-GOLD test set by uniformly sampling a smaller, balanced subset of criminal and civil cases. Three annotators independently extract and induce rule sets for this subset, following strict annotation guidelines. Details of the guidelines, annotator backgrounds, and inter-annotator agreement are provided in~\cref{app:annotation_guidelines}.

\subsection{Dataset Statistics and Expert Analysis}
\cref{tabs:dataset_stats} provides detailed LRI dataset statistics. The LRI-AUTO dataset comprises 5,121 case sets (totalling 38,088 cases and 29,650 rules), with 4,552 sets for training and 569 for testing.
The criminal-to-civil case ratio in LRI-AUTO (approx. 1:1.6) reflects the original CJO corpus distribution.
The LRI-GOLD test set contains 108 criminal and 108 civil case sets.
\cref{figs:heatmap} illustrates the numerical distribution of cases and rules per set across the dataset.
A manual audit conducted on 100 randomly selected LRI-AUTO sets, utilizing criteria specified in~\cref{tabs:data_quality_eval}, confirmed that the majority of rules correctly apply to their respective case sets.

\section{Experiments}\label{sec:experiments}

In this section, we assess LLMs performance on the LRI-GOLD benchmark and demonstrate how LRI-AUTO enhances legal rule induction in smaller models through parameter-efficient adaptation.

\begin{table*}[!t]
\renewcommand\arraystretch{1.00}
\small
\begin{center}
\begin{tabular}{
    m{2.2cm}|m{2.4cm}|
    m{1.2cm}<{\centering}m{1.2cm}<{\centering}|
    m{1.2cm}<{\centering}m{1.2cm}<{\centering}|
    m{1.2cm}<{\centering}m{1.2cm}<{\centering}
}
\toprule
\multirow{2}{*}{\textbf{Method}} & \multirow{2}{*}{\textbf{Model}} & 
\multicolumn{2}{c|}{\textbf{Rule Type}} & 
\multicolumn{2}{c|}{\textbf{Rule Level}} & 
\multicolumn{2}{c}{\textbf{Set Level}} \\
& & \textbf{Exp-Rec} & \textbf{Imp-Rec} & 
\textbf{Mic-Pre} & \textbf{Mic-F1} & 
\textbf{Mac-Pre} & \textbf{Mac-F1} \\
\midrule

\multirow{4}{*}{\makecell[l]{\textbf{LLMs}}}
& Llama-3.2-3B   & 24.91 & 9.97 & 19.13 & 19.21 & 17.99 & 17.89 \\
& Ministral-3B   & 41.49 & 21.14 & 36.54 & 35.18 & 37.01 & 32.83 \\
& Qwen-2.5-7B    & \textbf{58.23} & \textbf{33.73} & \textbf{56.45} & \textbf{52.53} & \textbf{57.97} & \textbf{50.75} \\
& Ministral-8B   & \underline{48.66} & \underline{21.38} & \underline{44.81} & \underline{41.43} & \underline{45.81} & \underline{39.78} \\
\midrule

\multirow{4}{*}{\makecell[l]{\textbf{LLMs + LoRA}}}
& Llama-3.2-3B & \underline{83.54} & 51.07 & \underline{70.47} & 70.96 & \underline{67.63} & 68.31 \\
& Ministral-3B & 78.96 & 38.05 & 60.79 & 62.25 & 55.61 & 58.48 \\
& Qwen-2.5-7B & \textbf{83.68} & \underline{56.06} & 70.07 & \underline{71.70} & 66.73 & \underline{68.65} \\
& Ministral-8B & 83.31 & \textbf{58.00} & \textbf{72.47} & \textbf{73.18} & \textbf{70.19} & \textbf{70.73} \\

\bottomrule
\end{tabular}
\end{center}
\caption{Performance~(\%) of four small-sized LLMs and their performance after LoRA fine-tuning on LRI-AUTO.}
\vspace{-0.15in}
\label{tabs:evaluation_finetune_performance}
\end{table*}

\subsection{Experimental Settings}

\paragraph{Baseline Methods}  
As discussed in~\cref{pre:pipeline}, we compare several approaches: Direct Induction (zero-shot prompting), CoT (prompting with "think step by step"), Long-CoT (reasoning before responding), SILVER (an automatic induction-verification pipeline), and fine-tuning on LRI-AUTO for small LLMs~(3B-8B).
Detailed prompt templates are provided in~\cref{app:implement}.

\paragraph{Models}  
We conduct experiments on three types of LLMs as depicted in~\cref{app:llm}: (1) LLMs: direct inference without thinking before response, (2) LRMs as detailed in~\cref{pre:LRM}, equipped with Long-CoT ability and think before response, (3) Small-size LLMs, whose parameter number is below or equal to 8 billion.

\paragraph{Evaluation Metrics}
We assess induced rule quality and correctness using DeepSeek-V3~\cite{deepseekai2025deepseekv3technicalreport} as an automated judge, employing two complementary perspectives:
(1) \textbf{Rule Level~(Micro) Evaluation}: This metric assesses all induced rules individually, disregarding their case set origins, to emphasize overall rule correctness (akin to micro-averaging). It is calculated as:
$
\small
\text{Mic-F1} = \frac{2 \cdot \text{Mic-Pre} \cdot \text{Mic-Rec}}{\text{Mic-Pre} + \text{Mic-Rec}},
$
where $\text{Mic-Pre}$ (micro-precision) is the total number of correctly predicted rules divided by the total number of predicted rules across all case sets, and $\text{Mic-Rec}$ (micro-recall) is the total number of correctly predicted rules divided by the total number of gold-standard rules across all case sets.
(2) \textbf{Set Level~(Macro) Evaluation}: This metric evaluates performance on a per-case-set basis, treating each as an independent unit and averaging their F1 scores:
$
\small
\text{Mac-F1} = \frac{1}{N_{\text{sets}}} \sum_{i=1}^{N_{\text{sets}}} \text{F1}(\mathcal{R}_i^{\text{pred}}, \mathcal{R}_i^{\text{gold}}),
$
where $\mathcal{R}_i^{\text{pred}}$ and $\mathcal{R}i^{\text{gold}}$ are the predicted and gold-standard rule sets for the $i$-th case set, and $N_{\text{sets}}$ is the total number of precedent case sets.
To assess the \textbf{reliability} of DeepSeek-V3 as an automated judge, we provide a detailed comparison with human evaluation results in~\cref{app:judge}.

\begin{figure}[t]
    \centering
    \includegraphics[width=0.45\textwidth]{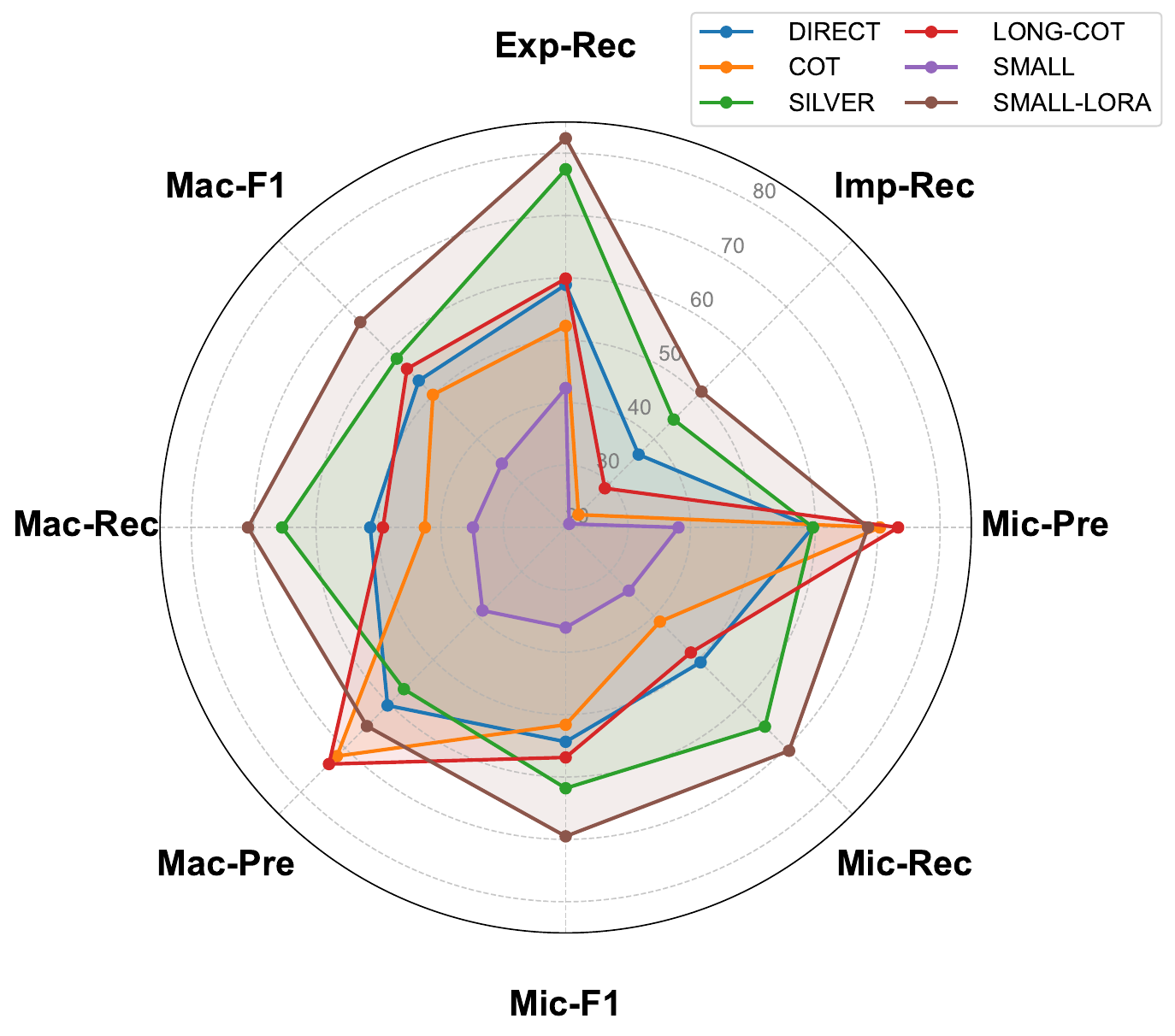}
    \caption{Scores (\%) of different baselines. For the Direct, CoT, and SILVER baselines, only the five LLMs common to all three are considered.}
    \label{figs:radar}
\end{figure}

\subsection{Main Evaluation}
\paragraph{Performance Comparison across Inductive Pipelines}\label{eval:preformance}
Analysis of \cref{tabs:evaluation_overall_performance} and \cref{figs:radar} reveals distinct performance characteristics of different inductive pipelines.
CoT prompting generally enhances precision at the cost of recall, leading to a slight decrease in F1 scores for most LLMs compared to Direct  Induction.
For instance, GPT-4o's~\cite{openai2024gpt4ocard} Micro-Precision rises from 71.81\% to 80.95\%, while its explicit rule recall drops from 55.56\% to 41.63\%.
Exceptions like Gemini-2.5-Flash (Mic-F1 +1.48\%) suggest model-specific benefits.
Long-CoT presents varied outcomes: Gemini-2.5-Flash~\cite{geminicard} (Long-CoT) improves precision (Mic-Pre +9.08\%) and Mic-F1 (+2.45\%) over its direct counterpart, albeit with reduced recall.
Conversely, Claude-3.7-Sonnet~\cite{Claude3S} (Long-CoT) showed increased recall (Exp-Rec +0.71\%) but lower precision (Mic-Pre -2.90\%) and Mic-F1 (-0.89\%).
This indicates that extended reasoning contexts affect the precision-recall balance differently across models.
The SILVER pipeline consistently yields superior performance, primarily through substantial recall improvements across models (\eg Gemini-2.5-Flash Exp-Rec increased from 73.00\% to 88.19\%).

\begin{figure*}[t]
    \centering
    \includegraphics[width=2\columnwidth]{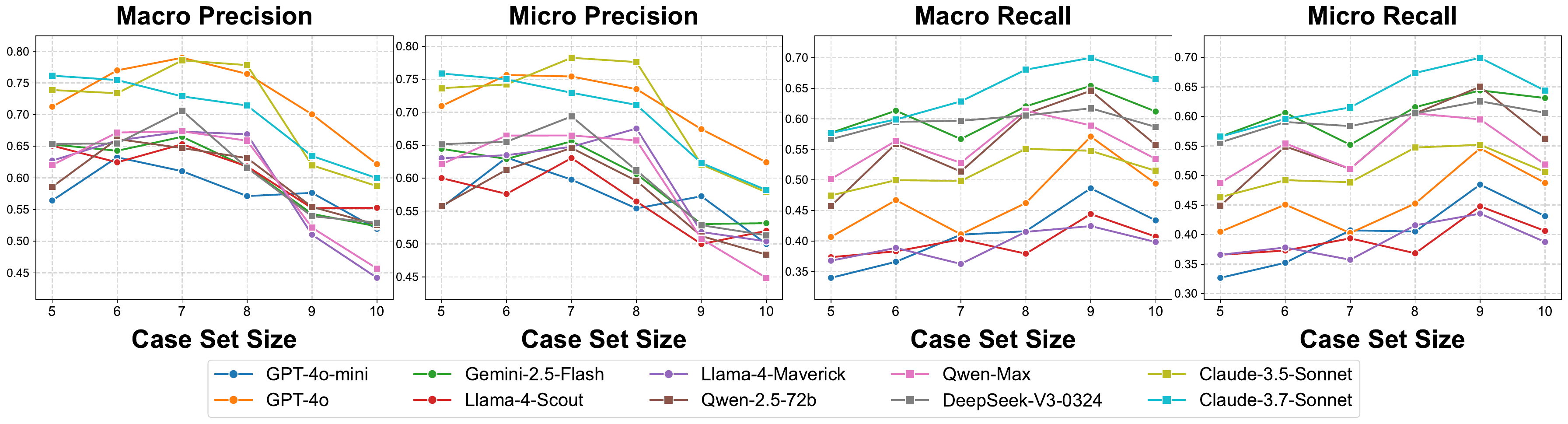}
    \caption{Performance trends of Direct Induction of ten LLMs across varying case set sizes.}
    \label{figs:evaluation_case_number}
\end{figure*}

\begin{figure}[!t]
    \centering
    \includegraphics[width=1\columnwidth]{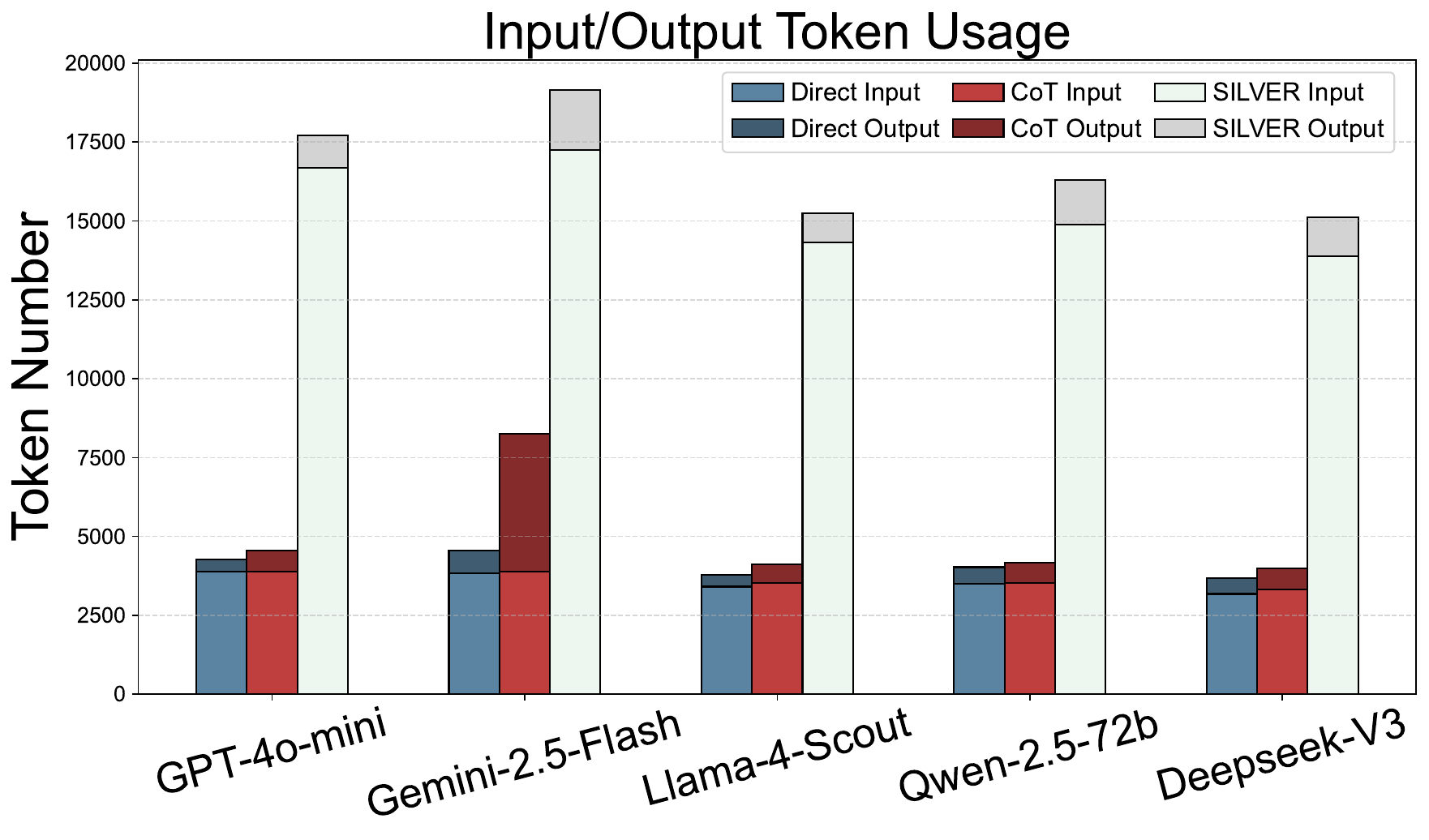}
    \caption{Comparison of token usage (Input \& Output) for different LLMs under three different baselines.}
    \label{figs:token_usage}
\end{figure}

\paragraph{Efficacy of LRI-AUTO}
\Cref{tabs:evaluation_finetune_performance} demonstrates the effectiveness of LRI-AUTO dataset in enhancing small LLMs (3B-8B) performance.
Initially, these models show limited capabilities (\eg Llama-3.2-3B Mic-F1 19.21\%).
However, LoRA fine-tuning~\cite{han2024parameterefficientfinetuninglargemodels} on LRI-AUTO yields substantial gains across all metrics for all four tested small LLMs. For example, Mic-F1 of Llama-3.2-3B surged to 70.96\%.
Notably, the fine-tuned Ministral-8B (+LoRA) achieves a Mic-F1 of 73.18\% and Mac-F1 of 70.73\%. This performance surpasses several larger proprietary models under Direct Induction prompting~(\cref{tabs:evaluation_overall_performance}), such as Gemini-2.5-Flash (Direct Mic-F1 60.51\%) and Claude-3.7-Sonnet (Direct Mic-F1 66.67\%).

\subsection{Further Discussion}
\paragraph{Explicit and Implicit Rule}
In the LRI evaluation phase, LLMs are not informed whether rules are explicit (directly from statutes) or implicit. We observe consistently higher recall for explicit rules.
We attribute this disparity to two primary factors. First, explicit rules are designed to be present across all cases within a given case set, which inherently increases their discoverability and ease of extraction by the models. Second, even when specific crime names are masked, LLMs with pre-existing knowledge of Chinese law (from their training data)~\cite{fei2023lawbenchbenchmarkinglegalknowledge} tend to exhibit greater sensitivity to the linguistic patterns characteristic of these explicit, statute-like rules.
Conversely, implicit rules, requiring deeper inference, are harder to identify. This suggests that performance on implicit rules may better reflect an LLM's ability to generalize in unfamiliar legal domains.

\paragraph{Set Size Sensitivity}\label{exp:sss}
As shown in~\cref{figs:evaluation_case_number}, LLM performance in legal rule induction varies with case set size. Generally, increasing the number of input cases leads to lower precision but higher recall. This is likely due to overgeneration of broad or less accurate rules, improving coverage (recall) but reducing accuracy (precision). With fewer cases, it's harder for models to detect shared patterns, leading to lower recall. Claude-3.7-Sonnet and DeepSeek-V3-0324 show stable performance across different sizes, while GPT-4o-mini and Llama-4-Maverick degrade more sharply, indicating difficulties in balancing abstraction and specificity.

\paragraph{Token Usage}
\cref{figs:token_usage} reveals that the SILVER pipeline incurs the highest token consumption due to its iterative multi-turn architecture.
Direct Induction prompting is the most token-efficient but, as noted in \cref{eval:preformance}, typically results in lower performance. The CoT strategy moderately increases token output compared to Direct Induction, but this often does not translate into commensurate F1 score improvements, potentially diminishing its cost-effectiveness. These observations underscore the trade-off between computational efficiency and reasoning depth in practical applications.

\section{Conclusion}
This paper formalizes Legal Rule Induction (LRI) as the task of distilling rules from analogous cases and introduces the first benchmark comprising LRI-AUTO for tuning and expert-annotated LRI-GOLD for evaluation. While situated in the legal domain, the core contribution extends beyond the extracted rules themselves: our experiments reveal that current LLMs struggle with inductive reasoning tasks, and demonstrate that targeted training on our dataset can substantially enhance this capability. This work establishes a foundation for LRI in the LLM era and addresses a critical gap in computational legal reasoning research.
\section*{Limitations}
Our research, conducted within the Chinese legal system, exclusively utilizes Chinese-language legal cases and rules. This grounding in a specific jurisdiction and language introduces limitations: the models may exhibit a bias towards the Chinese legal framework, potentially restricting their direct generalizability to other legal systems without adaptation, and their performance in multilingual contexts remains unassessed. Furthermore, this work did not investigate the utility of our legal rule induction methods on downstream applications such as legal information retrieval~\cite{SANSONE2022101967}, judgment prediction~\cite{cui2022surveylegaljudgmentprediction}, or question answering~\cite{MARTINEZGIL2023100552}; exploring this efficacy presents a significant avenue for future research.
A potential concern is pre-training data contamination, as court judgments used in our dataset may appear in LLM training corpora. We note two mitigating factors. First, since 2024, China Judgments Online has ceased releasing new court judgments, making it difficult to construct post-cutoff test data that would entirely eliminate this possibility. Second, and more importantly, the LRI task does not ask models to retrieve statutory provisions verbatim; rather, it requires synthesizing shared legal rules across analogous cases. These induced rules, particularly the implicit discretionary principles annotated by our legal experts, do not exist as such in any pre-training corpus, substantially reducing the risk of memorization-based shortcuts.

\section*{Ethics Statement}
The source materials for our dataset are derived from a publicly accessible platform, China Judgments Online, which is the largest legal judgment database in China and is widely used in academic research~\cite{huang-etal-2024-cmdl,DBLP:journals/corr/abs-1807-02478}.
Any specific legal provisions and personally identifiable information (PII) encountered are rigorously anonymised during the dataset construction process.
Human annotators involved in the project are compensated at a rate of 15 USD per hour, a figure that exceeds the prevailing minimum wage in China. To the best of our knowledge, this work adheres to all relevant open-source agreements and does not pose risks of information leakage or other ethical hazards.
Beyond data privacy and legality, we acknowledge that inducing legal rules from judicial texts may entail broader societal and ethical implications.
Due to the high-stakes nature of legal reasoning, the risk of LLM-induced hallucinations during rule induction demands careful scrutiny.
Models trained on judicial texts may reflect or amplify historical biases (e.g., related to gender or region) and could misrepresent complex legal reasoning if used without expert oversight. The LRI datasets and models are therefore released solely for academic research and methodological evaluation, with an emphasis on responsible use and further study of fairness, transparency, and accountability in computational law.

\section*{Acknowledgements}

The authors of this paper were supported by the National Key Research and Development Program of China (2025YFE0200500), the Beijing-Tianjin-Hebei Collaborative Innovation Special Project (26240201D), the ITSP Platform Research Project (ITS/189/23FP) from ITC of Hong Kong, SAR, China, and the AoE (AoE/E-601/24-N), the CRF (No. C6004-25G),  the RIF (R6021-20) and the GRF (16205322) from RGC of Hong Kong, SAR, China. The work described in this paper was conducted in full or in part by Dr. Haoran Li, JC STEM Early Career Research Fellow, supported by The Hong Kong Jockey Club Charities Trust.

\bibliography{custom}
\newpage
\appendix
\section{Legal Rule and Jurisprudence Foundation}\label{app:jur}
The foundational structure of a legal rule is commonly understood as an \textbf{if-then} conditional statement. This can be formally expressed in logical notation as:
\begin{equation}
    \textit{Condition} \rightarrow \textit{Consequence}
\end{equation}
In civil law jurisdictions, legal rules are typically explicitly stipulated and codified within statutes. For instance, specific articles within the \textit{Civil Code of the People's Republic of China}\footnote{\url{https://english.www.gov.cn/archive/lawsregulations/202012/31/content_WS5fedad98c6d0f72576943005.html}} or the French \textit{Code Civil (Napoleonic Code)}\footnote{\url{https://www.legifrance.gouv.fr/codes/texte_lc/LEGITEXT000006070721/}} clearly delineate such rules, providing a primary source for legal reasoning.
Conversely, common law rules are specific legal norms established by courts through precedent. Common law reasoning is also rule-based~\cite{Eisenberg_2022}, applying these court-derived rules to case facts. The rule a precedent establishes is its \textit{holding}—the explicit legal principle stated by the court as governing the case, which forms binding law. Other judicial statements within a precedent, known as \textit{dicta}, are not binding but may possess persuasive influence.
This paper, focusing on Chinese legal reasoning, adopts the ``three-element theory'' from Chinese jurisprudence. This theory structures a rule with a: \textbf{hypothetical condition}, \textbf{behavior pattern}, and \textbf{legal consequence} logically represented as:
\begin{equation}
\begin{aligned}
    \text{Hypothetical Condition} &\land \text{Behavior Pattern}\\&\rightarrow \text{Legal Consequence}
\end{aligned}
\end{equation}
An alternative, the new ``two-element theory'', posits rules as \textbf{constituent elements} and \textbf{legal consequences}. It suggests the behavior pattern is integrated within these two, aiming for a unified structure for various rule types.
However, we find that LLMs struggle to accurately interpret the new two-element theory, often producing erroneous outputs. Therefore, for reliability in this study, we utilize the more widely understood and LLM-compatible three-element theory.

\section{Details of LRI Dataset}
\subsection{China Judgments Online~(CJO)}\label{app:cjo}
This study utilizes CJO case data and legal article versions from 2021 due to two key factors. Primarily, the substantial public availability of 2021 case datasets makes them highly suitable for clustering purposes. Furthermore, the enactment of the Chinese Civil Code in 2020, which integrates seven distinct legal domains (General Provisions, Property Rights, Contracts, Personality Rights, Marriage and Family, Inheritance, and Tort Liability) previously governed by separate statutes, streamlines the process of systematic legal article extraction by avoiding the increased labor costs associated with mapping article numbers and content from a fragmented pre-2021 legal landscape.

\subsection{Prompt of Case Content Structuring}\label{app:prompts_ccs}

To facilitate a comprehensive presentation of a legal case's factual background, procedural history, judicial analysis, and adjudicated outcome, while concurrently ensuring the anonymisation of sensitive entities and the abstraction of specific legal article numbers and their textual content, we formulate the prompt delineated in~\cref{tabs:prompt_case_content_structuring}. This prompt utilizes the original legal case document and the content of cited legal provisions as input, which are subsequently processed by the DeepSeek-R1 model~\cite{deepseekai2025deepseekr1incentivizingreasoningcapability}.

\subsection{Prompt of Rule Extraction}\label{app:prompts_re}
Using the structured case contents and the content of cited legal provisions, we employ the prompts detailed in~\cref{tabs:prompt_rule_extraction} and~\cref{tabs:prompt_rule_extraction_continue}. Explicit and implicit rules are subsequently collected from the responses generated by DeepSeek-R1.

\subsection{Human Annotations}\label{app:annotation_guidelines}

To ensure high-quality rule extraction, we develop a concise and clear annotation guideline, as shown in~\cref{tabs:annotation_guideline}, based on the prompt designs in~\cref{tabs:prompt_rule_extraction} and~\cref{tabs:prompt_rule_extraction_continue}. The annotation task is carried out by three trained annotators, all graduate students in law from China, possessing a solid theoretical background in both Chinese criminal and civil law. Before commencing the full annotation process, we conduct a pilot annotation experiment to assess inter-annotator similarity. Specifically, the three annotators independently annotate 67 explicit and 39 implicit rules across 20 case sets. We then compute pairwise Jaccard similarities to evaluate the consistency of extracted rule sets among annotators. 
The results are presented in~\cref{tab:jaccard_similarity}. For explicit rules, the pairwise Jaccard similarities are relatively high (0.870–0.901), indicating strong agreement among annotators. In contrast, for implicit rules, the similarities are lower (0.553–0.574), suggesting that identifying implicit legal rules involves greater subjectivity and interpretive variation.

\begin{table}[t]
  \centering
  \resizebox{1\linewidth}{!}{%
  \begin{tabular}{lccc}
    \toprule
    & Annotator 1 & Annotator 2 & Annotator 3 \\
    \midrule
    \multicolumn{4}{l}{\textbf{Explicit Rules}} \\
    \midrule
    Annotator 1 & — & 0.870 & 0.887 \\
    Annotator 2 & 0.870 & — & 0.901 \\
    Annotator 3 & 0.887 & 0.901 & — \\
    \midrule
    \multicolumn{4}{l}{\textbf{Implicit Rules}} \\
    \midrule
    Annotator 1 & — & 0.553 & 0.571 \\
    Annotator 2 & 0.553 & — & 0.574 \\
    Annotator 3 & 0.571 & 0.574 & — \\
    \bottomrule
  \end{tabular}
  }
  \caption{Pairwise Jaccard similarities among annotators for explicit and implicit rules}
  \label{tab:jaccard_similarity}
\end{table}

\begin{table}[t]
\small
\centering
\centering
\begin{tabular}{p{0.96\columnwidth}}
\toprule
\begin{enumerate}
  \item \textbf{Legal Rule Categories:} Use only one of the following:
    (1) Criminal
    (2) Civil
    (3) Procedural (Litigation Procedure)

  \item \textbf{Legal Rule Structure:} Each rule must include:
  \begin{enumerate}
    \item \textbf{Hypothetical Condition} – the context and subject.
    \item \textbf{Behavior Pattern} – classified as:
    \begin{itemize}
      \item Permissive: “may”, “is allowed to”.
      \item Obligatory: “must”, “shall”.
      \item Prohibitive: “must not”, “is prohibited”.
    \end{itemize}
    \item \textbf{Legal Consequence} – result of compliance or violation.
  \end{enumerate}

  \item \textbf{Rule Types:}
  \begin{enumerate}
    \item \textbf{Explicit Rules:}
    \begin{itemize}
      \item Must be derived directly from cited laws.
      \item Must apply to \textbf{all} cases in the set.
      \item No article numbers or direct quotes.
    \end{itemize}
    \item \textbf{Implicit Rules:}
    \begin{itemize}
      \item Inferred from \textbf{majority}~(above 50\%) of cases.
      \item Reflect judicial discretion or practice.
    \end{itemize}
  \end{enumerate}

  \item \textbf{Formatting Requirements:}
  \begin{itemize}
    \item Follow the logic: \texttt{If [condition], and [behavior], then/otherwise [consequence]}.
    \item Avoid redundancy; merge similar rules.
    \item Avoid omissions; especially for cited laws.
    \item Replace legal terms with plain language.
  \end{itemize}

  \item \textbf{Metadata:} Count applicable cases for each rule.

  \item \textbf{Output Format:} Use JSON with keys: Explicit Rule, Implicit Rule.
\end{enumerate}
\\
\bottomrule
\end{tabular}
\caption{Annotation guideline for legal rule induction}
\label{tabs:annotation_guideline}
\end{table}

\subsection{Statistics}\label{app:statistics}
\begin{figure*}[t]
    \centering
    \includegraphics[width=2\columnwidth]{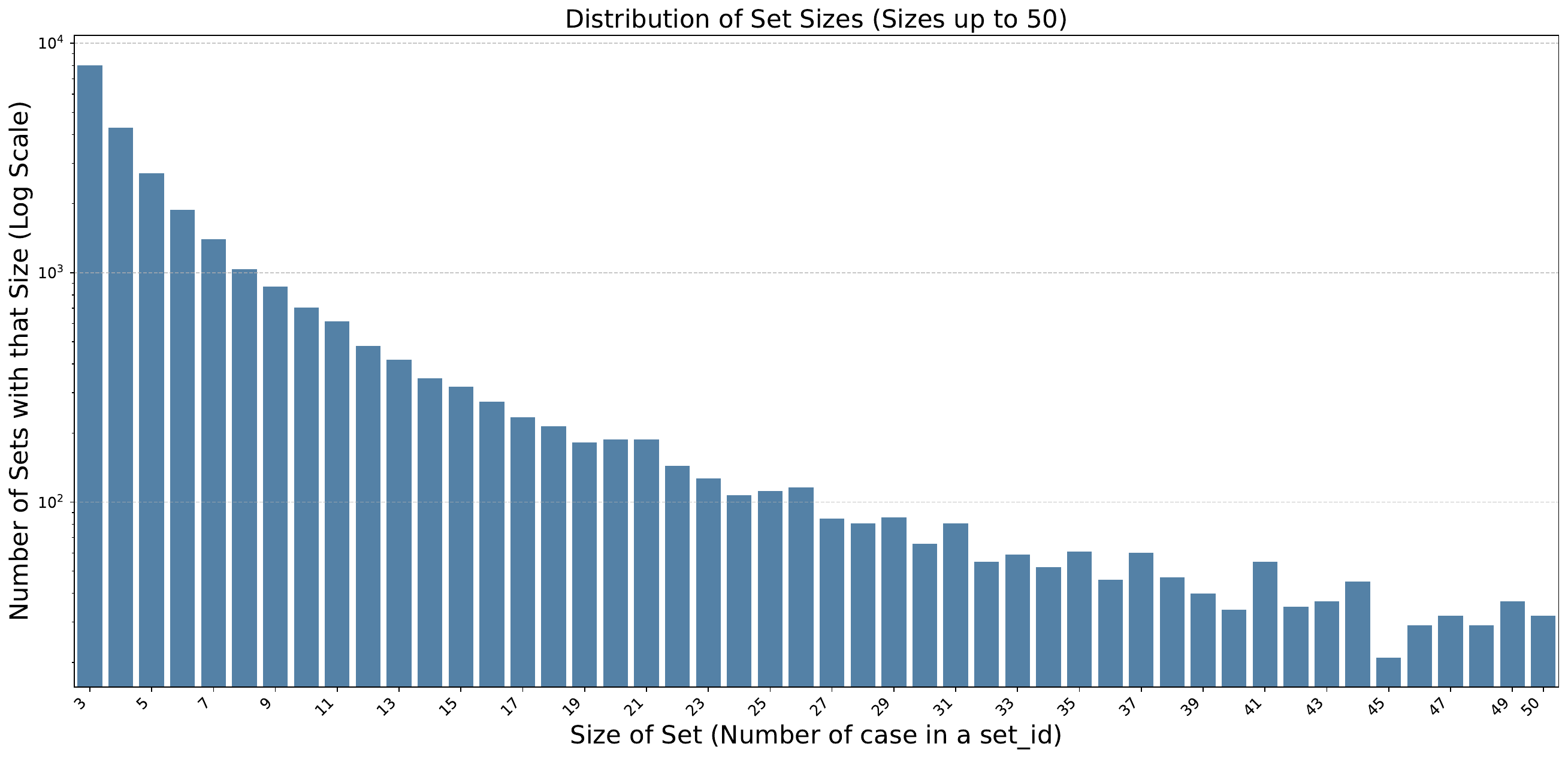}
    \caption{Original case set size distribution before re-sampling.}
    \label{figs:set_size_distribution}
    \vspace{-0.15in}
\end{figure*}

\begin{figure}[t]
    \centering
    \includegraphics[width=1\columnwidth]{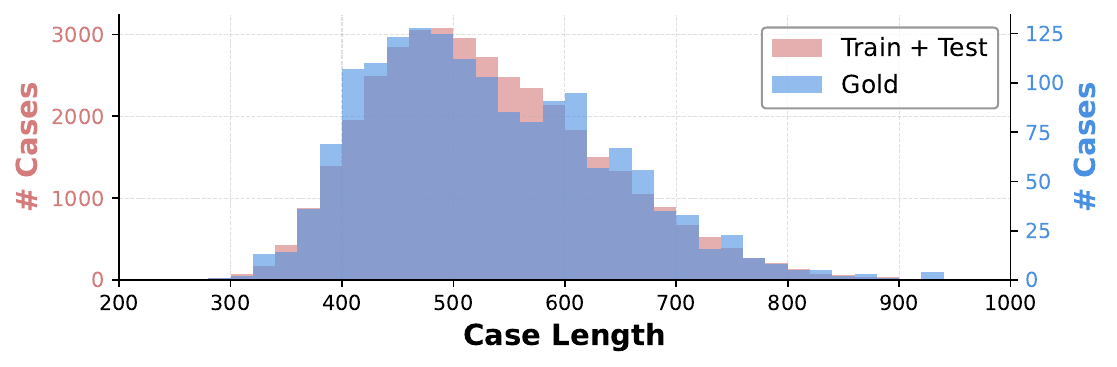}
    \caption{Case length distribution in LRI dataset.}
    \label{figs:case_length_distribution}
    \vspace{-0.15in}
\end{figure}

\begin{figure}[t]
    \centering
    \includegraphics[width=1\columnwidth]{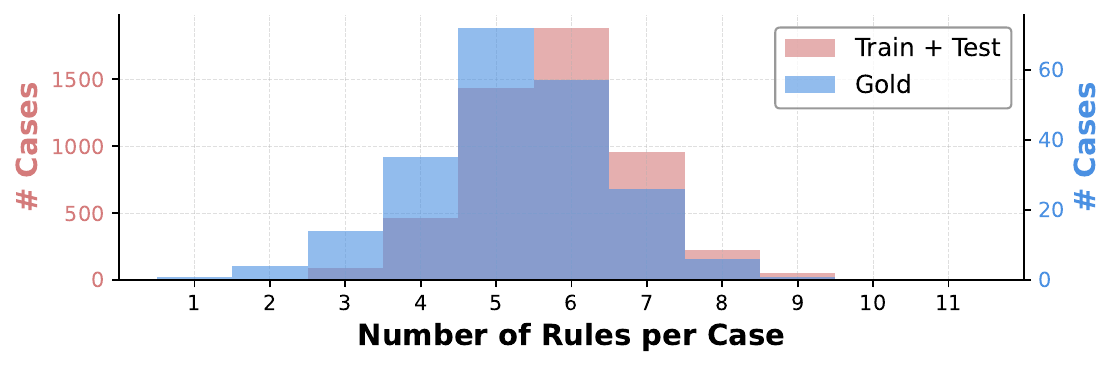}
    \caption{Rule number per case set distribution in LRI dataset.}
    \label{figs:rule_number_distribution}
    \vspace{-0.15in}
\end{figure}

\cref{figs:set_size_distribution} illustrates the original case set size distribution. ~\cref{figs:case_length_distribution} depicts the case length distribution within the LRI dataset, with most cases ranging from 400 to 600 Chinese characters in length. Furthermore, ~\cref{figs:rule_number_distribution} presents the distribution of rules per case set in the LRI dataset. We observe that the number of rules per case set in LRI-GOLD is slightly lower than in LRI-AUTO.

\section{Implementation Details}~\label{app:implement}
\subsection{Model Details}\label{app:llm}
In our experiments, we evaluate a total of 19 LLMs, categorized as follows:
\paragraph{LLMs} GPT-4o-mini~\cite{openai2024gpt4ocard}, GPT-4o~\cite{openai2024gpt4ocard}, Gemini-2.5-Flash~\cite{geminicard}, Llama-4-Scout~\cite{llama4herd}, Llama-4-Maverick~\cite{llama4herd}, Qwen-2.5-72b~\cite{qwen2025qwen25technicalreport}, Qwen-Max~\cite{qwen2025qwen25technicalreport}, DeepSeek-V3-0324~\cite{deepseekai2025deepseekv3technicalreport}, Claude-3.5-Sonnet~\cite{Claude35}, Claude-3.7-Sonnet~\cite{Claude3S}.
\paragraph{LRMs} DeepSeek-R1~\cite{deepseekai2025deepseekr1incentivizingreasoningcapability}, o3-mini~\cite{o3mini}, Grok-3-mini~\cite{grok3}, Claude-3.7-Sonnet:Thinking~\cite{Claude3S}, Gemini-2.5-Flash:Thinking~\cite{geminicard}.
\paragraph{Small LLMs} Llama-3.2-3B~\cite{llama32}, Ministral-3B~\cite{ministral3b}/Ministral-8B~\cite{ministral8b}, Qwen-2.5-7B~\cite{qwen2025qwen25technicalreport}.

All LLMs and LRMs are accessed via the OpenRouter API\footnote{\url{https://openrouter.ai/}}, while the small LLMs are obtained from Hugging Face\footnote{\url{https://huggingface.co/}}.

\begin{algorithm}[t]
\caption{The pipeline of simply iterative induction and verification}
\begin{algorithmic}[1]
\Require Case set $\mathcal{P}$, Threshold $\tau$ (\eg 50\%), Maximum iterations $\text{max\_iter}$
\Ensure Final rule set $\mathcal{R}_{\text{final}}$
\State $\mathcal{R}_{\text{final}} \gets \emptyset$
\State $\mathcal{R}_{\text{cand}} \gets$ \Call{InduceInitialRules}{$\mathcal{P}$}
\State $iter \gets 0$
\While{$iter < \text{max\_iter}$}
    \State $\mathcal{R}_{\text{verified}} \gets$ \Call{VerifyAndSelect}{$\mathcal{R}_{\text{cand}}$, $\mathcal{P}$, $\tau$}
    \If{$\mathcal{R}_{\text{verified}} = \emptyset$}
        \State \textbf{break}
    \EndIf
    \State $\mathcal{R}_{\text{final}} \gets \mathcal{R}_{\text{final}} \cup \mathcal{R}_{\text{verified}}$
    \State $\mathcal{R}_{\text{cand}} \gets$ \Call{InduceNewRules}{$\mathcal{P}$, $\mathcal{R}_{\text{final}}$}
    \State $iter \gets iter + 1$
\EndWhile
\State \Return $\mathcal{R}_{\text{final}}$
\end{algorithmic}
\label{alg:silver}
\end{algorithm}

\subsection{LoRA Training Setting}\label{app:lora}
Four open-source language models, each supporting at least an 8K token input context, are selected for instruction-tuning on the GPU with 80GB of VRAM and 1,513 TFLOPS. Specifically, these models are fine-tuned using Low-Rank Adaptation~(LoRA)~\cite{hu2021loralowrankadaptationlarge} for parameter efficiency. For LoRA, both the rank and alpha are set to 8. All models are trained for 3 epochs, and their final checkpoints are used for evaluation. Other training parameters include a batch size of 8, a learning rate of 1e-4, a cutoff length of 8192 tokens, and a warmup ratio of 0.1. The training time for each model ranges from 4 to 8 hours.

\subsection{LLM-as-a-Judge}\label{app:judge}
Given that a single legal rule can be expressed in various linguistic forms, standard automatic evaluation metrics such as exact match, ROUGE, and BLEU are unsuitable for assessing rule induction. Consequently, we used DeepSeek-V3 as an LLM-as-a-Judge to evaluate the logical equivalence between induced rules and ground-truth rules. The prompts utilized for this evaluation are detailed in~\cref{tabs:prompt_eval_rule_quality}.
To check the quality of the LLM-as-a-Judge, we manually reviewed the judgments made by DeepSeek-V3 on 114 rules. These rules are sampled by selecting 3 rules from each of the 38 distinct models and settings from the test phase. The outcomes of this quality assessment, presented in~\cref{tabs:judge_quality_eval}, show that DeepSeek-V3 performs with high accuracy on this type of classification task.

\subsection{Prompt of SILVER}\label{app:prompts_silver}
The SILVER workflow includes three main stages. First, an initial round of legal rule induction is performed using the prompt specified in~\cref{tabs:prompt_rule_extraction_eval}. Second, a legal rule verification step, utilizing the prompt in~\cref{tabs:prompt_rule_verify}, checks if each induced rule applies to a majority (over 50\%) of cases within the given case set. Third, a subsequent round of inducing new rules from the case set is conducted, guided by the prompts detailed in~\cref{tabs:prompt_induce_new_rules} and~\cref{tabs:prompt_induce_new_rules_continue}. This stage uses the legal case set and the rule set generated in the preceding round as input.

\subsection{Prompt of Direct Induction (Evaluation Phase)}\label{app:prompts_lri_eval}
For the evaluation phase of legal rule induction, we design the prompt shown in~\cref{tabs:prompt_rule_extraction_eval} and~\cref{tabs:prompt_rule_extraction_eval_continue}. The input for this prompt consists solely of the legal case set, without any cited legal provisions. It is employed with both LLMs~(Direct) and LRMs.

\section{Supplementary Experimental Results}

\subsection{Set Size Sensitivity (Supplement)}\label{app:sss_sup}
\begin{figure*}[t]
    \centering
    \includegraphics[width=2\columnwidth]{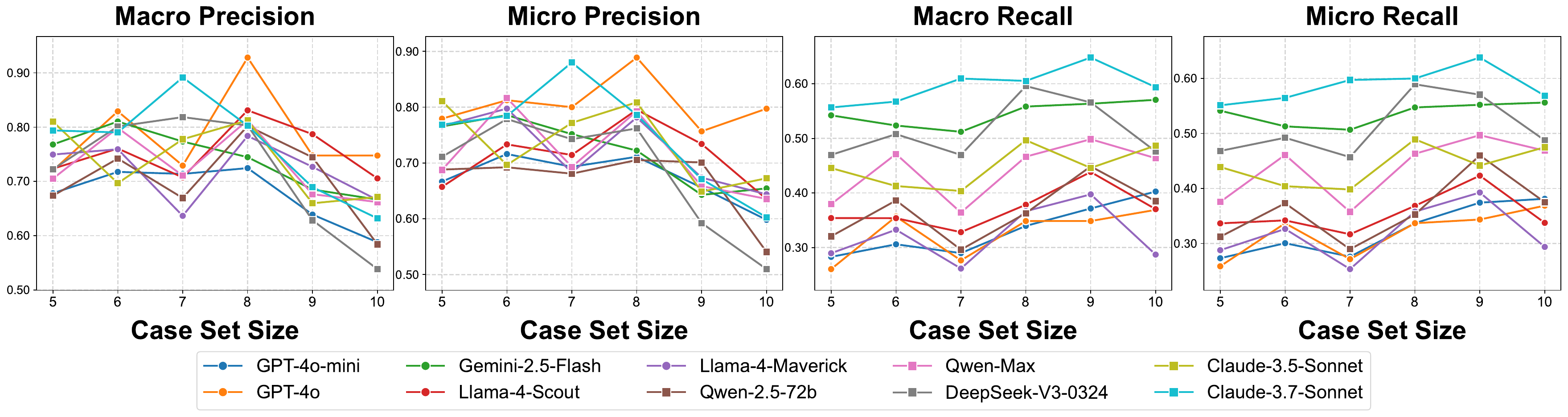}
    \caption{Performance trends of CoT of ten LLMs across varying case set sizes.}
    \label{figs:evaluation_case_number_cot}
    \vspace{-0.15in}
\end{figure*}

\begin{figure*}[t]
    \centering
    \includegraphics[width=2\columnwidth]{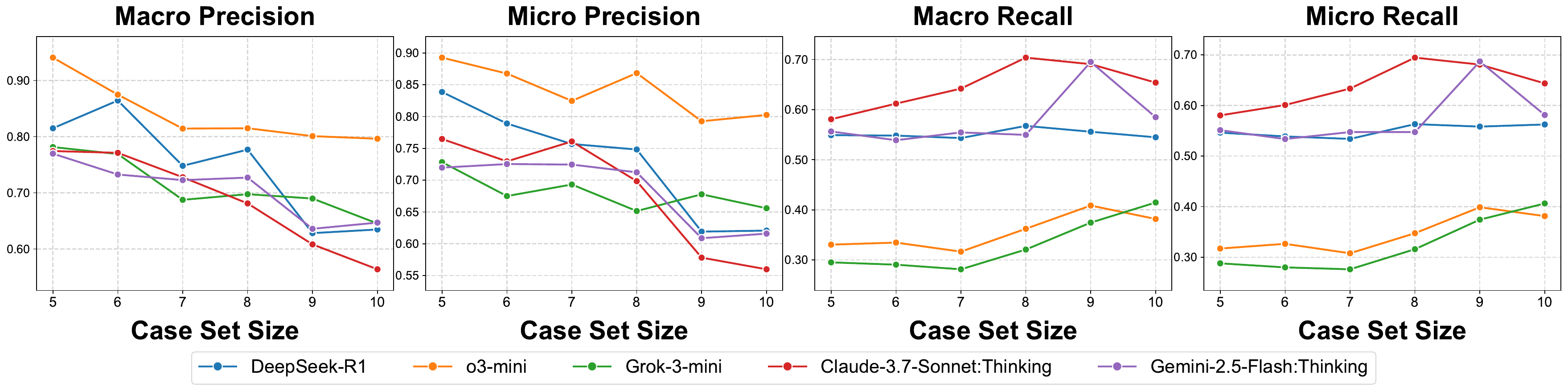}
    \caption{Performance trends of Long-CoT of five LRMs across varying case set sizes.}
    \label{figs:evaluation_case_number_lrm}
    \vspace{-0.15in}
\end{figure*}

\begin{figure*}[t]
    \centering
    \includegraphics[width=2\columnwidth]{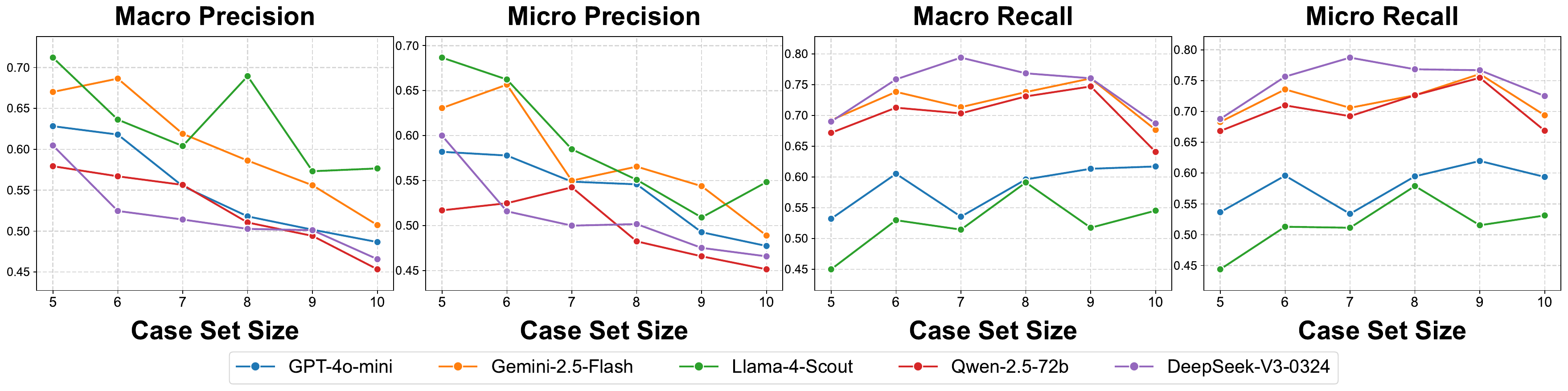}
    \caption{Performance trends of SILVER of five LLMs across varying case set sizes.}
    \label{figs:evaluation_case_number_workflow}
    \vspace{-0.15in}
\end{figure*}

This section presents additional data on set size sensitivity for other inductive pipelines: CoT (~\cref{figs:evaluation_case_number_cot}), Long-CoT (~\cref{figs:evaluation_case_number_lrm}), and SILVER (~\cref{figs:evaluation_case_number_workflow}). These results further support the conclusions drawn in~\cref{exp:sss}.

\subsection{Case Study}\label{app:case_study}
To illustrate the structure and quality of cases within our dataset, we present representative examples of both a criminal case (\cref{figs:criminal_case_study}) and a civil case (\cref{figs:civil_case_study}). Each case is processed using the full case-processing pipeline introduced in~\cref{sec:rule_extraction}.

\subsection{Error Analysis}
To further assess the effect of fine-tuning on legal rule induction, we perform a comparative analysis of inference outputs from the \texttt{Llama-3.2-3B} model before and after training on the LRI-AUTO dataset. The comparison, shown in \cref{figs:llama_case_study}, serves as a focused \textit{error analysis} illustrating how fine-tuning reduces hallucinations, aligns induced rules with ground-truth logic, and improves the accuracy of legal terminology and structural coherence.
Before fine-tuning, the \texttt{Llama-3.2-3B} model exhibits several systematic deficiencies:
\begin{itemize}
  \item \textbf{Semantic Drift:} The model frequently hallucinates domain-irrelevant concepts (e.g., tax obligations or traffic negligence) that are absent from the case facts. This reflects an inability to constrain generation to the domain-specific legal frame.
  \item \textbf{Structure Inconsistency:} Generated rules often deviate from the tripartite logical form (\textit{Hypothetical Condition}–\textit{Behavior Pattern}–\textit{Legal Consequence}), merging components or omitting legal outcomes.
  \item \textbf{Terminological Hallucination:} The pre-trained model uses incorrect or generic legal terms (e.g., “increase responsibility” rather than “sentence aggravation”), diminishing alignment with professional terminology.
\end{itemize}

After fine-tuning on the LRI-AUTO dataset, these issues are largely mitigated:
\begin{itemize}
  \item Generated rule structures adhere closely to the expected tripartite form.
  \item Hallucinations and off-topic insertions drop sharply. Induced rules align with actual case fact patterns.
  \item Implicit rules (e.g., voluntary surrender, compensation) emerge with appropriate conditional logic and proportionate legal consequences.
  \item Both procedural and substantive law distinctions are captured, with consistent and context-sensitive vocabulary.
\end{itemize}

The comparative analysis (\cref{figs:llama_case_study}) evidences a substantial shift from generic, loosely structured outputs to logically coherent, legally valid formulations. This improvement supports the hypothesis that large-scale LRI datasets can significantly enhance inductive capacity in LLMs.


\begin{table*}[t]
\setlength\doublerulesep{\arrayrulewidth}
\renewcommand\cellset{\renewcommand\arraystretch{1}}
\centering
\begin{tabular}{lc}
\toprule
Judge Quality Assessment Question & Yes \% \\
\midrule
\makecell[l]{Is the assessment of element completeness correct?}&100.0\% \\ \midrule
\makecell[l]{Is the assessment of sensitive content correct?}&100.0\% \\ \midrule
\makecell[l]{Is the assessment of rule coverage correct?}&98.24\% \\ \midrule\midrule
Is the final assessment conclusion correct?& 97.36\% \\
\bottomrule
\end{tabular}
\caption{Human analysis of DeepSeek-V3 judge quality.}
\label{tabs:judge_quality_eval}
\end{table*}

\begin{table*}[t]
\setlength\doublerulesep{\arrayrulewidth}
\renewcommand\cellset{\renewcommand\arraystretch{1}}
\centering
\begin{tabular}{lc}
\toprule
Rule Quality Review Question & Yes (\%) \\
\midrule
\makecell[l]{Is the explicit rule applicable to all the cases in its set?} & 94.21\% \\
\midrule
\makecell[l]{Is the implicit rule applicable to more than half of the cases in its set?} & 95.03\% \\
\midrule
\makecell[l]{Is the rule logically consistent and does it use legal terminology appropriately?} & 99.59\% \\
\midrule
\makecell[l]{Is the rule distinct and not redundant with other rules?} & 100.0\% \\
\midrule\midrule
Are all fields in this rule correct? & 93.63\% \\
\bottomrule
\end{tabular}
\caption{Human evaluation of LRI-AUTO data quality.}
\label{tabs:data_quality_eval}
\end{table*}

\begin{table*}[t!]

\small
\centering
\begin{tabular}{{p{0.9\textwidth}}}
\toprule
A legal case typically includes a description of the facts, legal analysis, relevant legal provisions, and the ruling.
Please read the given legal case and extract the following four parts: Fact Description, Litigation Process, Legal Analysis, and Judgment Result.
\\
\par 

\textbf{[Element Definitions]} \par
\textbf{Fact Description:} The basic circumstances of the case and the core dispute (maintaining the integrity of the events). \par
\textbf{Litigation Process:} The trial process and procedural matters of the case. \par
\textbf{Legal Analysis:} The reasoning process of the judgment (reflecting the logic of legal application). \par
\textbf{Judgment Result:} The final disposition and conclusion. \par

\par 
\\
Please process the following case: \par
\textbf{\{Legal Case\}} \par
The legal provisions cited in this case are as follows: \par
\textbf{\{Legal Provisions\}} \par
\\
\par 

\textbf{[Extraction Rules]} \par
\textbf{(1) Content Requirements}
\begin{itemize}
    \itemsep0pt 
    \item Prioritize using the original wording; key details must not be omitted. Do not summarize; do not summarize; do not summarize.
    \item The legal analysis must reflect the logic of how the provisions were applied, but specific article numbers/content of the provisions should not appear.
    \item Direct citation of charges or legal terms is prohibited (e.g., use "caused property loss to others" instead of "theft"). This is especially true for the Judgment Result section.
    \item There should be no redundant information or logical contradictions among the four parts.
    \item The four parts should be able to corroborate each other and the cited legal provisions, reflecting the application logic of the legal provisions.
\end{itemize}

\vspace{0.5ex} 
\textbf{(2) Desensitization Norms}
\begin{itemize}
    \itemsep0pt 
    \item Replace all entities with pseudonyms (People: A/B/C; Organizations: Company A/Unit B; Locations: Place C).
    \item Basic identity information such as gender, age, and occupation should be retained.
    \item Remove court information (replace specific court names with "adjudicating authority"); remove personal information of judges, lawyers, etc.
\end{itemize}

\vspace{0.5ex} 
\textbf{(3) Output Format} \par
Output according to the following JSON format:
\begin{verbatim}
{
    "Fact Description": "XXX",
    "Litigation Process": "XXX",
    "Legal Analysis": "XXX",
    "Judgment Result": "XXX"
}
\end{verbatim}
\\

\bottomrule
\end{tabular}
\caption{Prompt of case content structuring.}
\label{tabs:prompt_case_content_structuring}
\end{table*}
\begin{table*}[t!]

\small
\centering
\begin{tabular}{{p{0.9\textwidth}}}
\toprule

Please extract legal rules from the following set of legal cases and the corresponding legal provisions, and output in the required format.

\par 

\textbf{[Element Definitions]} \par
Each legal rule must contain the following three components: \par
\textbf{1. Hypothetical Conditions:} Conditions and circumstances under which the rule applies, including applicable subjects and their behaviors. \par
\textbf{2. Behavioral Pattern:} Specifies how people should act, including permissive, obligatory, and prohibitive patterns. \par
- \textbf{Permissive pattern:} Uses expressions such as “may,” “is entitled to,” or “is allowed to.” \par
- \textbf{Obligatory pattern:} Uses expressions such as “shall,” “must,” or “has the obligation to.” \par
- \textbf{Prohibitive pattern:} Uses expressions such as “prohibited,” “shall not,” or “must not.” \par
\textbf{3. Legal Consequence:} Specifies the consequences of complying or not complying with the behavioral pattern. \par
- Positive consequence: Legal effect resulting from compliance. \par
- Negative consequence: Legal liability resulting from violation. \par

\par

Here are examples of the three behavioral patterns: \par
1. \textbf{Permissive:} \par
Hypothetical Condition: A natural person wishes to engage in a civil transaction. \par
Behavioral Pattern: The person may (but is not required to) enter into a contract. \par
Legal Consequence: If a contract is formed, the person is bound by it; if not, there is no contractual obligation. \par

2. \textbf{Obligatory:} \par
Hypothetical Condition: Citizens, legal persons, or other organizations meet the conditions for tax liability (e.g., taxable income). \par
Behavioral Pattern: Must pay taxes on time and in full. \par
Legal Consequence: If taxes are paid lawfully, rights are enjoyed normally; if not, there may be fines, late fees, or other liabilities. \par

3. \textbf{Prohibitive:} \par
Hypothetical Condition: A natural person with full criminal responsibility. \par
Behavioral Pattern: Prohibited from committing theft. \par
Legal Consequence: If no theft is committed, there is no liability; if theft occurs, the person may face criminal penalties, such as detention, fines, or imprisonment. \par

\vspace{0.5ex}
\textbf{[Extraction Rules]} \par
\textbf{I. Explicit Rule Extraction} \par
\begin{itemize}
    \item Must directly correspond to the cited legal provisions and reflect their core content;
    \item May combine multiple relevant provisions into a composite rule;
    \item Direct reference to specific article numbers or content is prohibited; instead, summarize into a general rule applicable to the case set;
    \item Explicit rules must apply to all cases in the set.
\end{itemize}

\vspace{0.5ex}
\textbf{II. Implicit Rule Extraction} \par
\begin{itemize}
    \item Must be inferred from commonalities among cases and not directly derived from legal provisions;
    \item Should reflect discretionary standards in judicial practice;
    \item Must apply to most cases in the set (i.e., more than half).
\end{itemize}

\vspace{0.5ex}
Example: From all traffic accident cases, infer that “if the driver fails to exercise reasonable care, liability may be increased.” \par

\vspace{0.5ex}
\textbf{III. General Requirements} \par
\begin{itemize}
    \item Each rule must include all three components to form a complete logical chain: “If [Hypothetical condition], then [behavioral pattern], and [legal consequence] follows.”
    \item Type must be one of: Criminal / Civil / Procedural; do not use other types.
    \item Avoid duplication; merge similar rules.
    \item Do not omit rules, especially those clearly reflected in the cited legal provisions.
    \item Do not use legal terminology or charges directly (e.g., use “caused property loss to others” instead of “theft”).
\end{itemize}
\\

\bottomrule
\end{tabular}

\caption{Prompt of legal rule extraction from case set.}
\label{tabs:prompt_rule_extraction}
\end{table*}

\begin{table*}[t!]

\small
\centering
\begin{tabular}{{p{0.9\textwidth}}}
\toprule

\vspace{0.5ex}
The following is the set of legal cases: \par
\textbf{\{Legal Case Set\}} \par

The legal provisions cited in the case set are as follows: \par
\textbf{\{Legal Provisions\}} \par
\\
\vspace{0.5ex}
\textbf{[Output Format]} \par
Please output in the following JSON format: \par
\begin{verbatim}
{
  "Explicit Rules": [
    {
      "Applicable Case Count": 10,
      "Type": "Criminal",
      "Content": {
        "Hypothetical Condition": "A natural person with full criminal responsibility",
        "Behavioral Pattern": {
          "Type": "Prohibitive",
          "Description": "Prohibited from intentionally and unlawfully depriving others of life"
        },
        "Legal Consequence": "If a person kills, they may face the death penalty, life imprisonment,
        or fixed-term imprisonment of over ten years"
      }
    }
  ],
  "Implicit Rules": [
    {
      "Applicable Case Count": 10,
      "Type": "Criminal",
      "Content": {
        "Hypothetical Condition": "The suspect has voluntarily surrendered",
        "Behavioral Pattern": {
          "Type": "Obligatory",
          "Description": "Should truthfully confess the main facts of the offense"
        },
        "Legal Consequence": "May receive a lighter or mitigated punishment according to law"
      }
    }
  ],
  "Unreflected Provisions": {
    "Civil Code of the People’s Republic of China": ["Article 111"],
    "Civil Procedure Law of the People’s Republic of China": ["Article 120", "Article 131"]
  }
}
\end{verbatim}
\\
\bottomrule
\end{tabular}

\caption{Prompt of legal rule extraction from case set. (Continue)}
\label{tabs:prompt_rule_extraction_continue}

\end{table*}

\begin{table*}[t!]
\small
\centering
\begin{tabular}{{p{0.9\textwidth}}}
\toprule

Please extract legal rules from the following set of legal cases and output in the required format.

\par 

\textbf{[Element Definitions]} \par
Each legal rule must contain the following three components: \par
\begin{itemize}
\item \textbf{1. Hypothetical Conditions:} The part of a legal rule concerning the conditions and circumstances for its application, including conditions for application and conditions for the subject's behavior. \par
\item \textbf{2. Behavioral Pattern:} The part of a legal rule that specifies how people should act, including permissive (authorization) patterns, obligatory (duty) patterns, and prohibitive (prohibition) patterns. \par
  - \textbf{Permissive pattern:} Uses authorizing expressions such as “may,” “is entitled to,” or “is allowed to.” \par
  - \textbf{Obligatory pattern:} Uses mandatory expressions such as “shall,” “must,” or “has the obligation to.” \par
  - \textbf{Prohibitive pattern:} Uses prohibitive expressions such as “prohibited,” “shall not,” or “must not.” \par
\item \textbf{3. Legal Consequence:} The part of a legal rule that specifies the corresponding results people should bear when their actions comply with or violate the requirements of the behavioral pattern. \par
  - \textbf{Positive consequence:} The legal effect resulting from compliance with the behavioral pattern. \par
  - \textbf{Negative consequence:} The legal liability resulting from violation of the behavioral pattern. \par
\end{itemize}
\par
Here are examples of legal rules for the three behavioral patterns: \par
\begin{itemize}
\item \textbf{1. Permissive:} \par
Hypothetical Condition: A natural person wishes to engage in a civil transaction. \par
Behavioral Pattern: The natural person may (but is not required to) enter into a contract. \par
Legal Consequence: If a contract is entered into, they are legally bound by the contract; if no contract is entered into, there is no contractual obligation. \par

\item \textbf{2. Obligatory:} \par
Hypothetical Condition: Citizens, legal persons, and other organizations meet the conditions for tax liability (e.g., have taxable income). \par
Behavioral Pattern: Must pay taxes on time and in full. \par
Legal Consequence: If taxes are paid according to law, rights are enjoyed normally; if taxes are not paid according to law, they may face fines, late fees, or other legal liabilities. \par

\item \textbf{3. Prohibitive:} \par
Hypothetical Condition: A natural person with full criminal responsibility. \par
Behavioral Pattern: Prohibited from committing theft. \par
Legal Consequence: If no theft is committed, there is no legal liability; if theft is committed, they may face criminal penalties, such as detention, fines, or fixed-term imprisonment. \par
\end{itemize}
\vspace{0.5ex}
\textbf{[Extraction Rules]} \par
1. Each rule must include all three components, forming a complete logical chain: \par
   “If [Hypothetical condition], then [behavioral pattern], then/otherwise [legal consequence].” \par
2. Do not use specific article numbers, content, or charges; summarize into a general rule applicable to the given case set. \par
3. Must be inferred from commonalities among cases and should reflect discretionary standards in judicial practice. \par
4. The extracted rules must apply to $\geq$ 51\% of the cases. \par 
   Example: Infer from all traffic accident cases in the set that “if the driver fails to exercise reasonable care, liability may be increased.” \par
5. Combining multiple relevant provisions to form a composite rule is allowed. \par
6. Type annotation: Criminal / Civil / Procedural; do not use other types. \par
7. Avoid duplication; merge similar rules. \par
\\

\bottomrule
\end{tabular}
\caption{Prompt of legal rule induction from a case set in the evaluation phase.}
\label{tabs:prompt_rule_extraction_eval}
\end{table*}

\begin{table*}[t!]
\small
\centering
\begin{tabular}{{p{0.9\textwidth}}}
\toprule

\vspace{0.5ex}
The set of legal cases is as follows: \par
\textbf{\{Legal Case Set\}} \par
\\
\vspace{0.5ex}
\textbf{[Output Format]} \par
Please output in the following JSON format: \par
\begin{verbatim}
{
  "Extracted Rules": [
    {
      "Type": "Criminal",
      "Content": {
        "Hypothetical Condition": "A natural person with full criminal responsibility",
        "Behavioral Pattern": {
          "Type": "Prohibitive",
          "Description": "Prohibited from intentionally and unlawfully depriving others of life"
        },
        "Legal Consequence": "If a person kills, they face the death penalty, life imprisonment,
        or fixed-term imprisonment of over ten years"
      }
    },
    {
      "Type": "Procedural",
      "Content": {
        "Hypothetical Condition": "The plaintiff in a civil case files a lawsuit",
        "Behavioral Pattern": {
          "Type": "Obligatory",
          "Description": "Shall provide clear claims and factual reasons when filing the lawsuit"
        },
        "Legal Consequence": "If the requirements are met, the case shall be accepted;
        if the requirements are not met, a one-time notice for correction shall be given"
      }
    }
  ]
}
\end{verbatim}
\\
\bottomrule
\end{tabular}
\caption{Prompt for legal rule induction from case set in the evaluation phase (Continue).}
\label{tabs:prompt_rule_extraction_eval_continue}
\end{table*}
\begin{table*}[t!]
\small
\centering
\begin{tabular}{{p{0.9\textwidth}}}
\toprule

Please extract legal rules from the following set of legal cases and output in the required format.

\par 

\textbf{[Element Definitions]} \par
Each legal rule must contain the following three components: \par
\begin{itemize}
    \item \textbf{1. Hypothetical Conditions:} The part of a legal rule concerning the conditions and circumstances for its application, including conditions for application and conditions for the subject's behavior. \par
    \item \textbf{2. Behavioral Pattern:} The part of a legal rule that specifies how people should act, including permissive (authorization) patterns, obligatory (duty) patterns, and prohibitive (prohibition) patterns. \par
      - \textbf{Permissive pattern:} Uses authorizing expressions such as “may,” “is entitled to,” or “is allowed to.” \par
      - \textbf{Obligatory pattern:} Uses mandatory expressions such as “shall,” “must,” or “has the obligation to.” \par
      - \textbf{Prohibitive pattern:} Uses prohibitive expressions such as “prohibited,” “shall not,” or “must not.” \par
    \item \textbf{3. Legal Consequence:} The part of a legal rule that specifies the corresponding results people should bear when their actions comply with or violate the requirements of the behavioral pattern. \par
        - \textbf{Positive consequence:} The legal effect resulting from compliance with the behavioral pattern. \par
        - \textbf{Negative consequence:} The legal liability resulting from violation of the behavioral pattern. \par
\end{itemize}
\par
Here are examples of legal rules for the three behavioral patterns: \par
\begin{itemize}
\item \textbf{1. Permissive:} \par
Hypothetical Condition: A natural person wishes to engage in a civil transaction. \par
Behavioral Pattern: The natural person may (but is not required to) enter into a contract. \par
Legal Consequence: If a contract is entered into, they are legally bound by the contract; if no contract is entered into, there is no contractual obligation. \par

\item \textbf{2. Obligatory:} \par
Hypothetical Condition: Citizens, legal persons, and other organizations meet the conditions for tax liability (e.g., have taxable income). \par
Behavioral Pattern: Must pay taxes on time and in full. \par
Legal Consequence: If taxes are paid according to law, rights are enjoyed normally; if taxes are not paid according to law, they may face fines, late fees, or other legal liabilities. \par

\item \textbf{3. Prohibitive:} \par
Hypothetical Condition: A natural person with full criminal responsibility. \par
Behavioral Pattern: Prohibited from committing theft. \par
Legal Consequence: If no theft is committed, there is no legal liability; if theft is committed, they may face criminal penalties, such as detention, fines, or fixed-term imprisonment. \par
\end{itemize}
\vspace{0.5ex}
\textbf{[Extraction Rules]} \par
1. Each rule must include all three components, forming a complete logical chain: \par
   “If [Hypothetical condition], and [behavioral pattern], then/otherwise [legal consequence].” \par
2. Do not use specific article numbers, content, or charges; summarize into a general rule applicable to the given case set. \par
3. Must be inferred from commonalities among cases and should reflect discretionary standards in judicial practice. \par
4. The extracted rules must apply to $\geq$51\% of the cases. \par
   Example: Infer from all traffic accident cases in the set that “if the driver fails to exercise reasonable care, liability may be increased.” \par
5. Combining multiple relevant provisions to form a composite rule is allowed. \par
6. Type annotation: Criminal / Civil / Procedural; do not use other types. \par
7. Avoid duplication; merge similar rules. \par
\\
\bottomrule
\end{tabular}
\caption{Prompt of new rule induction.}
\label{tabs:prompt_induce_new_rules}
\end{table*}

\begin{table*}[t!]
\small
\centering
\begin{tabular}{{p{0.9\textwidth}}}
\toprule

\vspace{0.5ex}
The set of legal cases is as follows: \par
\{Legal Case Set\} \par

\vspace{0.5ex}
The rules already extracted are as follows, please do not extract them again: \par
\{Already Extracted Rules\} \par
Please do not extract existing rules again to avoid redundancy. \par

\vspace{0.5ex}
\textbf{[Output Format]} \par
Please output in the following JSON format: \par
\begin{verbatim}
{
  "Extracted Rules": [
    {
      "Type": "Criminal",
      "Content": {
        "Hypothetical Condition": "A natural person with full criminal responsibility",
        "Behavioral Pattern": {
          "Type": "Prohibitive",
          "Description": "Prohibited from intentionally and unlawfully depriving others of life"
        },
        "Legal Consequence": "If a person kills, they face the death penalty, life imprisonment,
        or fixed-term imprisonment of over ten years"
      }
    },
    {
      "Type": "Procedural",
      "Content": {
        "Hypothetical Condition": "The plaintiff in a civil case files a lawsuit",
        "Behavioral Pattern": {
          "Type": "Obligatory",
          "Description": "Shall provide clear claims and factual reasons when filing the lawsuit"
        },
        "Legal Consequence": "If the requirements are met, the case shall be accepted;
        if the requirements are not met, a one-time notice for correction shall be given"
      }
    }
  ]
}
\end{verbatim}
\\
\bottomrule 
\end{tabular}
\caption{Prompt of new rule induction (Continue).}
\label{tabs:prompt_induce_new_rules_continue}
\end{table*}
\begin{table*}[t!]
\small
\centering
\begin{tabular}{{p{0.9\textwidth}}}
\toprule

Please verify the applicable case count and structural integrity of the following legal rules based on the given set of legal cases and legal rules.
\\
\par 

\textbf{[Element Definitions]} \par
Each legal rule must contain the following three components: \par
\begin{itemize}
\item \textbf{1. Hypothetical Conditions:} The part of a legal rule concerning the conditions and circumstances for its application, including conditions for application and conditions for the subject's behavior. \par
\item \textbf{2. Behavioral Pattern:} The part of a legal rule that specifies how people should act, including permissive (authorization) patterns, obligatory (duty) patterns, and prohibitive (prohibition) patterns. \par
  - \textbf{Permissive pattern:} Uses authorizing expressions such as “may,” “is entitled to,” or “is allowed to.” \par
  - \textbf{Obligatory pattern:} Uses mandatory expressions such as “shall,” “must,” or “has the obligation to.” \par
  - \textbf{Prohibitive pattern:} Uses prohibitive expressions such as “prohibited,” “shall not,” or “must not.” \par
\item \textbf{3. Legal Consequence:} The part of a legal rule that specifies the corresponding results people should bear when their actions comply with or violate the requirements of the behavioral pattern. \par
  - \textbf{Positive consequence:} The legal effect resulting from compliance with the behavioral pattern. \par
  - \textbf{Negative consequence:} The legal liability resulting from violation of the behavioral pattern. \par
\end{itemize}
\vspace{0.5ex}
The set of legal cases is as follows: \par
\textbf{\{Legal Case Set\}} \par

\vspace{0.5ex}
The rule set to be evaluated is as follows: \par
\textbf{\{Rule Set to be Evaluated\}} \par
\\
\vspace{0.5ex}
Output only the JSON-formatted content; do not add any explanatory text. \par
\textbf{[Output Format]} \par
\begin{verbatim}
{
  "Evaluation Results": [
    { "Rule ID": 1, "Applicable Case Count": 10 (Assumed value, should be calculated 
    based on the case set), "Rule Integrity": "Complete"/"Incomplete"},
    { "Rule ID": 2, "Applicable Case Count": 7 (Assumed value, should be calculated 
    based on the case set), "Rule Integrity": "Complete"/"Incomplete"},
    { "Rule ID": 3, "Applicable Case Count": ...
    
  ]
}
\end{verbatim}
\\
\bottomrule
\end{tabular}
\caption{Prompt for legal rule verification.}
\label{tabs:prompt_rule_verify}
\end{table*}

\begin{table*}[t!]
\small
\centering
\begin{tabular}{p{0.95\textwidth}}
\toprule
\textbf{Multi-dimensional assessment of target rules based on a legal rule quality assessment framework.} \\

\vspace{1ex}

\textbf{[Assessment Object]} \\
\textbf{Rule to be assessed:} \\
\texttt{\{Rule to be assessed\}} \\

\textbf{Reference Rule Sets:} \\
\textbf{Explicit Rule Set (directly corresponding to legal articles):} \\
\texttt{\{Explicit rule set\}} \\
\textbf{Implicit Rule Set (judicial practice conventions):} \\
\texttt{\{Implicit rule set\}} \\

\vspace{1ex}

\textbf{[Assessment Criteria]} \\

\textbf{1. Three-element check}
\begin{itemize}
    \item \textbf{Hypothetical Condition:} Whether the preconditions for rule application are clearly defined.
    \item \textbf{Behavioral Pattern:} Whether the type (may do/should do/must not do) is accurately marked and described.
    \item \textbf{Legal consequences:} Whether it includes the positive and negative consequences corresponding to the Behavioral Pattern.
\end{itemize}

\textbf{2. Prohibited content check}
\begin{itemize}
    \item Whether there are prohibited references such as legal article numbers, names of crimes, etc.
\end{itemize}

\textbf{3. Rule coverage check}
\begin{itemize}
    \item Whether it is logically equivalent to any rule in the explicit rule set.
    \item Whether it is logically equivalent to any rule in the implicit rule set.
\end{itemize}

\textbf{4. Assessment conclusion}
\begin{itemize}
    \item Rules that meet all the above requirements are \textbf{"Correct"}.
    \item In the coverage check, \textbf{"logical equivalence"} must be achieved to be considered \textbf{"Correct"}.
    \item If it does not meet the three-element check or contains prohibited content, it is \textbf{"Incorrect"}.
    \item If it does not match any explicit or implicit rules, it is \textbf{"Incorrect"}.
\end{itemize}

\vspace{1ex}

\textbf{[Output Format]} \\

\begin{verbatim}
{
  "Element Completeness": {
    "Hypothetical Condition": "Not Present"/"Correct"/"Incorrect",
    "Behavioral Pattern": "Not Present"/"Correct"/"Incorrect",
    "Legal Consequences": "Not Present"/"Correct"/"Incorrect"
  },
  "Prohibited/Sensitive Content": "Present"/"Not Present",
  "Rule Coverage": {
    "Explicit Rules": "Logically Equivalent"/"Partially Matches"/"Does Not Match",
    "Implicit Rules": "Logically Equivalent"/"Partially Matches"/"Does Not Match"
  },
  "Assessment Conclusion": "Correct"/"Incorrect"
}
\end{verbatim}
\\
\bottomrule
\end{tabular}
\caption{Prompt of multi-dimensional assessment of legal rules.}
\label{tabs:prompt_eval_rule_quality}
\end{table*}

\begin{figure*}[t]
\begin{tcolorbox}[
  colback = cBlue_1!5,
  colframe = cBlue_6,
  coltitle = white,
  fonttitle = \bfseries\small,
  fontupper = \small,
  fontlower = \small,
  title = {§ Rule Set Comparison (Ground-Truth vs Induced)}
]

\textbf{§ Ground-Truth Rule Set:}

\begin{itemize}
  \item \textbf{Explicit Rule 1 (Criminal)}  
    \textit{Hypothetical Condition:} A natural person with full criminal responsibility commits an act endangering public safety.
    \textit{Behavior Pattern:} Prohibition of endangering public safety by arson, explosion, or other dangerous methods.
    \textit{Legal Consequence:} 3–10 years imprisonment; leniency for voluntary surrender, guilty plea, etc.

  \item \textbf{Explicit Rule 2 (Litigation Procedure)}  
    \textit{Hypothetical Condition:} Suspect pleads guilty and accepts punishment, no procedural flaws.  
    \textit{Behavior Pattern:} Charges and sentencing recommendations should be adopted.  
    \textit{Legal Consequence:} Court adopts recommendations unless manifestly improper.

  \item \textbf{Implicit Rule 1 (Criminal)}  
    \textit{Hypothetical Condition:} Voluntary surrender and truthful confession.  
    \textit{Behavior Pattern:} Should cooperate with judicial proceedings.  
    \textit{Legal Consequence:} Leniency (20–40\% sentence reduction).

  \item \textbf{Implicit Rule 2 (Criminal)}  
    \textit{Hypothetical Condition:} Compensation for losses after endangering public safety.  
    \textit{Behavior Pattern:} Should obtain victim's forgiveness.  
    \textit{Legal Consequence:} Discretionary leniency (10–30\% sentence reduction).

  \item \textbf{Implicit Rule 3 (Criminal)}  
    \textit{Hypothetical Condition:} Multiple overlapping dangers.  
    \textit{Behavior Pattern:} Prohibition of dangerous acts in flammable areas.  
    \textit{Legal Consequence:} Base sentence increased to mid-to-upper range.
\end{itemize}

\vspace{0.5em}
\textbf{§ Induced Rule Set Before Training:}

\begin{itemize}
  \item \textbf{Rule 1 (Explicit/Criminal)}  
    \textit{Hypothetical Condition:} Driver failed to exercise reasonable care.  
    \textit{Behavior Pattern:} Prohibition of inattentive driving.  
    \textit{Legal Consequence:} May increase responsibility.

  \item \textbf{Rule 2 (Explicit/Criminal)}  
    \textit{Hypothetical Condition:} Tax obligations apply.  
    \textit{Behavior Pattern:} Must pay taxes in full and on time.  
    \textit{Legal Consequence:} Legal rights if paid; fines or liabilities if not.

  \item \textbf{Rule 3 (Explicit/Criminal)}  
    \textit{Hypothetical Condition:} Natural person with criminal responsibility.  
    \textit{Behavior Pattern:} Prohibition of theft.  
    \textit{Legal Consequence:} Punishment if theft committed.
\end{itemize}

\vspace{0.5em}
\textbf{§ Induced Rule Set After Training:}

\begin{itemize}
  \item \textbf{Rule 1 (Explicit/Criminal)}  
    \textit{Hypothetical Condition:} Criminal act endangering public safety.  
    \textit{Behavior Pattern:} Prohibition of arson or similar acts.  
    \textit{Legal Consequence:} Up to 3 years imprisonment; 3–10 years for serious cases.

  \item \textbf{Rule 2 (Explicit/Litigation Procedure)}  
    \textit{Hypothetical Condition:} Guilty plea and clear facts.  
    \textit{Behavior Pattern:} Sentencing recommendations should be adopted.  
    \textit{Legal Consequence:} Judgment consistent with recommendations.

  \item \textbf{Rule 3 (Implicit/Criminal)}  
    \textit{Hypothetical Condition:} Voluntary surrender or confession.  
    \textit{Behavior Pattern:} Truthfully confess main facts.  
    \textit{Legal Consequence:} Leniency or mitigation possible.

  \item \textbf{Rule 4 (Implicit/Criminal)}  
    \textit{Hypothetical Condition:} Compensation and forgiveness.  
    \textit{Behavior Pattern:} Fulfill civil compensation obligations.  
    \textit{Legal Consequence:} Discretionary leniency.

  \item \textbf{Rule 5 (Explicit/Criminal)}  
    \textit{Hypothetical Condition:} Surrender or guilty plea.  
    \textit{Behavior Pattern:} Sign plea agreement and accept sentencing.  
    \textit{Legal Consequence:} Lenient procedures applied.
\end{itemize}

\end{tcolorbox}
\caption{An example of a rule set generated by \textbf{Llama-3.2-3B} before and after training on the LRI-AUTO dataset.}
\label{figs:llama_case_study}
\end{figure*}

\begin{figure*}[t]
\begin{tcolorbox}[
  colback = cBlue_1!5,
  colframe = cBlue_6,
  coltitle = white,
  fonttitle = \bfseries\small,
  fontupper = \small,
  fontlower = \small,
  title = {§ A Criminal Case Example}
]
\textbf{Fact Description:}
The defendant (male, born in 1978, with a junior high school education, and working as a crew member), due to a personal dispute with Person A, set fire to dry straw in the bedroom of the house he shared with Person A at around 11:00 p.m. on February 18, 2020, after consuming alcohol. He also recorded a video of the act and sent it to Person A. The house is located in Area C and was rented by Person A. It is adjacent to Person B's residence on the west side, 1.3 meters from Person C's residence on the east, and across the street from Person D's house to the south. There was a haystack beside the street.\\
\textbf{Litigation Process:}
The case was prosecuted by the public prosecution authority and publicly tried by a lawfully formed collegial panel. The prosecution alleged that the defendant’s actions constituted a crime of endangering public safety, presenting evidence such as victim statements, witness testimonies, and on-site inspection records. The defendant and his defense counsel did not dispute the charges. The defense argued for leniency based on voluntary surrender and admission of guilt. The trial court confirmed eight categories of evidence presented and challenged during the hearing.\\
\textbf{Legal Analysis:}
The court determined that the defendant intentionally committed arson by setting fire to another person's property, which posed a substantial danger to public safety. Although the act did not result in severe consequences, the fire occurred in a densely populated area with flammable materials nearby, presenting a real risk. The defendant voluntarily turned himself in and truthfully confessed, which constitutes a legal ground for leniency. He also voluntarily admitted guilt and accepted punishment, qualifying for a more lenient sentence. The sentencing recommendation by the prosecution was deemed appropriate given the facts and circumstances and was adopted by the court.\\
\textbf{Judgment Result:}
The defendant was sentenced to three years and six months of fixed-term imprisonment, with the sentence commencing on February 19, 2020. The court considered mitigating factors such as voluntary surrender, truthful confession, and admission of guilt when determining the sentence. The time already spent in detention was credited toward the prison term.
\end{tcolorbox}
\caption{A criminal case from CJO after case processing.}
\label{figs:criminal_case_study}
\end{figure*}

\begin{figure*}[t]
\begin{tcolorbox}[
  colback = cBlue_1!5,
  colframe = cBlue_6,
  coltitle = white,
  fonttitle = \bfseries\small,
  fontupper = \small,
  fontlower = \small,
  title = {§ A Civil Case Example}
]
\textbf{Fact Description:}
On March 18, 2019, Party A (male, born August 11, 1969, Han ethnicity) applied for a loan through his electronic banking account with Bank A, signing the “Quick e-Loan Agreement” and the “Loan Service Agreement” electronically (via data message).
The contract stipulated a loan amount of 71,500 RMB, with a term from March 18, 2019, to March 18, 2020, and an annual interest rate of 5.6\%. In case of overdue payments, the penalty interest rate would increase by 50\%.
Bank A disbursed the loan as agreed, but Party A failed to make repayments according to the contract.\\
\textbf{Litigation Process:}
The case was filed on April 15, 2021. The court applied summary procedures and held a public hearing on May 25, 2021.
Bank A’s authorized litigation representative attended the trial. Party A, though legally summoned, did not appear in court, so the court conducted a trial in absentia.\\
\textbf{Legal Analysis:}
The loan agreements signed electronically by both parties reflected their true intent and contained legally valid content, making the contracts legally binding and effective.
Since Bank A fulfilled its obligation by disbursing the loan, and Party A breached the agreement by failing to repay, he is liable to return the principal and pay the agreed interest and penalty interest.
As for Bank A’s claims for announcement and asset preservation fees, the court did not support them due to a lack of evidence proving that those expenses are actually incurred.\\
\textbf{Judgment Result:}
Party A is ordered to repay Bank A the loan principal of 71,500 RMB within ten days after the judgment takes effect, along with interest and penalty interest as stipulated in the contract.
Other claims made by Bank A are dismissed.
If Party A fails to fulfill the monetary obligations on time, it must pay double interest on the overdue amount during the delay period.
The case acceptance fee of 790 RMB is to be borne by Party A.\\
\end{tcolorbox}
\caption{A civil case from CJO after case processing.}
\label{figs:civil_case_study}
\end{figure*}
\end{document}